\documentclass[format=acmsmall, review=false, screen=true]{acmart}
\usepackage{animate} 

\begin{document}

\title[Smart Director: An Event-Driven Directing System for Live Broadcasting]{Smart Director: An Event-Driven Directing System\\ for Live Broadcasting}

\author{Yingwei~Pan}
\affiliation{%
  \institution{JD AI Research}
  \city{Beijing}
  \country{China}}
\email{panyw.ustc@gmail.com}
\author{Yue~Chen}
\affiliation{%
  \institution{JD AI Research}
  \city{Beijing}
  \country{China}}
\email{chenyue21@jd.com}
\author{Qian~Bao}
\affiliation{%
  \institution{JD AI Research}
  \city{Beijing}
  \country{China}}
\email{baoqian@jd.com}
\author{Ning~Zhang}
\affiliation{%
  \institution{JD AI Research}
  \city{Mountain View}
  \country{USA}}
\email{ning.zhang@jd.com}
\author{Ting~Yao}
\affiliation{%
  \institution{JD AI Research}
  \city{Beijing}
  \country{China}}
\email{tingyao.ustc@gmail.com}
\author{Jingen~Liu}
\affiliation{%
  \institution{JD AI Research}
  \city{Mountain View}
  \country{USA}}
\email{jingenliu@gmail.com}
\author{Tao~Mei}
\affiliation{%
  \institution{JD AI Research}
  \city{Beijing}
  \country{China}}
\email{tmei@jd.com}


\begin{abstract}
Live video broadcasting normally requires a multitude of skills and expertise with domain knowledge to enable multi-camera productions. As the number of cameras keep increasing, directing a live sports broadcast has now become more complicated and challenging than ever before. The broadcast directors need to be much more concentrated, responsive, and knowledgeable, during the production. To relieve the directors from their intensive efforts, we develop an innovative automated sports broadcast directing system, called Smart Director, which aims at mimicking the typical human-in-the-loop broadcasting process to automatically create near-professional broadcasting programs in real-time by using a set of advanced multi-view video analysis algorithms. Inspired by the so-called ``three-event'' construction of sports broadcast \cite{goldlust1987playing}, we build our system with an event-driven pipeline consisting of three consecutive novel components: 1) the \textit{Multi-view Event Localization} to detect events by modeling multi-view correlations, 2) the \textit{Multi-view Highlight Detection} to rank camera views by the visual importance for view selection, 3) the \textit{Auto-Broadcasting Scheduler} to control the production of broadcasting videos. To our best knowledge, our system is the first end-to-end automated directing system for multi-camera sports broadcasting, completely driven by the semantic understanding of sports events. It is also the first system to solve the novel problem of multi-view joint event detection by cross-view relation modeling. We conduct both objective and subjective evaluations on a real-world multi-camera soccer dataset, which demonstrate the quality of our auto-generated videos is comparable to that of the human-directed. Thanks to its faster response, our system is able to capture more fast-passing and short-duration events which are usually missed by human directors.
\end{abstract}

\setcopyright{acmcopyright}
\acmJournal{TOMM}
\acmYear{2021} \acmVolume{1} \acmNumber{1} \acmArticle{1} \acmMonth{1} \acmPrice{15.00}\acmDOI{10.1145/3448981}

%
%




\maketitle

\renewcommand{\shortauthors}{Y. Pan et al.}
\section{Introduction}
\begin{figure}[!tb]
\vspace{-0.0in}
	\centering\includegraphics[width=0.78\textwidth]{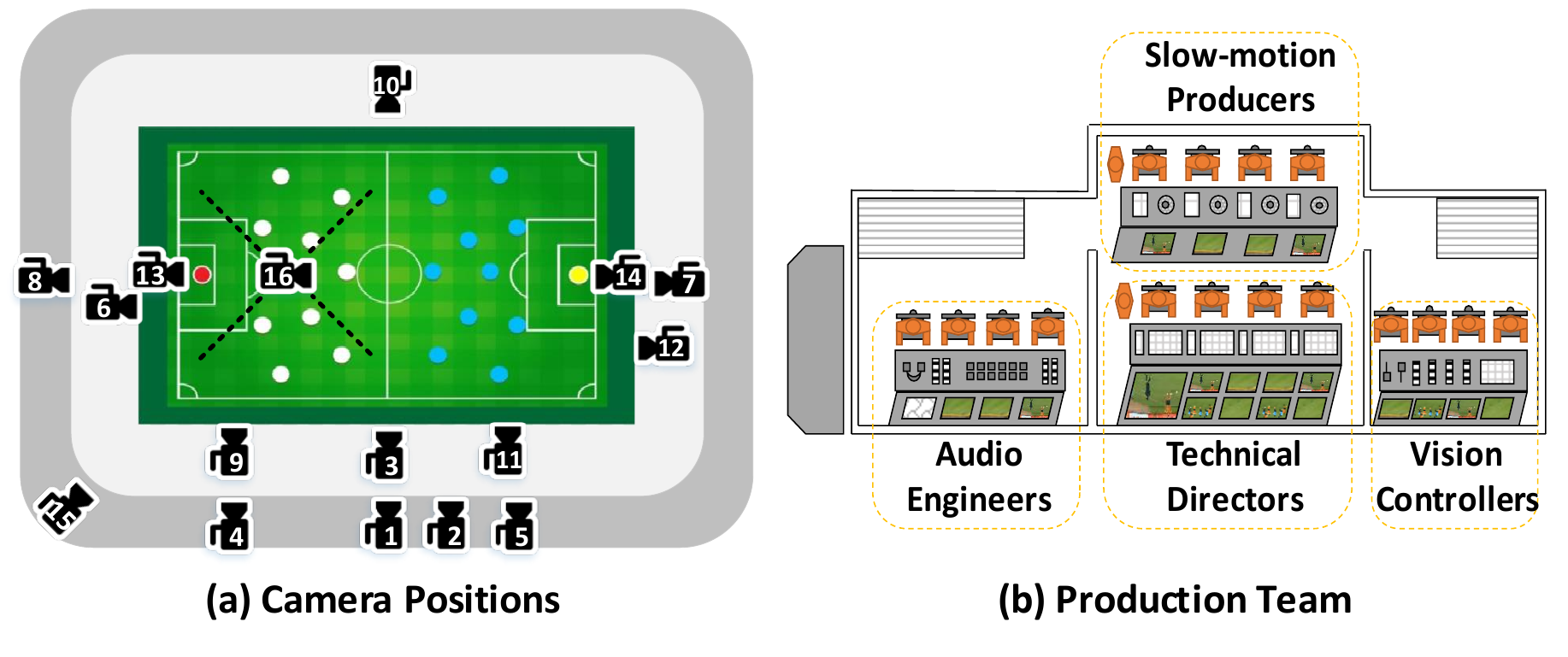}
    \vspace{-0.2in}
	\caption{\small To complete a broadcasting for a typical soccer match, the number of cameras placed in a variety of places of a stadium like (a) can be 15+, and the production team like (b) typically consists of 30+ members to perform various operations including directing, making slow-motion replays, visual graphics, and camera switch. The core goal of our work is to develop a smart system to mimic the human directing process by leveraging its capability of event understanding and learning-based auto-broadcasting scheduling.}
\label{figintro}
\vspace{-0.10in}
\end{figure}

Sports broadcasting is a storytelling process \cite{owens2015television}, during which the director leverages multiple cameras placed in a variety of places (e.g., Figure \ref{figintro} (a)) throughout the stadium to bring the viewers an immersive feeling in the game. In a soccer broadcast, for example, the director typically uses a main camera to cover the general action of the game, a ``hero camera'' to take the close-ups of outstanding efforts, and some ``special-assignment'' cameras to cover specific action shots, such as ``foul'' and ``offside'' \cite{owens2015television}. During the broadcasting, the directors play a crucial role in the production team (e.g., Figure \ref{figintro} (b) shows a typical production team layout in an outside broadcasting vehicle), who are in charge of directing cameras, slow-motion replays and visual graphics. In other words, the directors' responsibility is to determine when and how to switch camera views and what content to put on-air to satisfy the viewers' entertainment pleasure. With the increasing number of cameras used for broadcasting (e.g., about 33 cameras for FIFA 2018 WorldCup \footnote{\url{https://football-technology.fifa.com/en/innovations/var-at-the-world-cup/}}), however, the live-broadcast directing is getting more challenging and overwhelming for human directors. Additionally, as the broadcasting bandwidth is increasing in the 5G era, the viewers are expected to be entertained with the personalized broadcasting \cite{wang2016personal} to satisfy their preference. Obviously, it is infeasible for human directors to accomplish such huge broadcasting workloads. Therefore, in this paper we are dedicated to developing an automated directing system, which can automatically produce broadcast videos.

\begin{figure*}[!tb]
\vspace{-0.00in}
	\centering\includegraphics[width=0.96\textwidth]{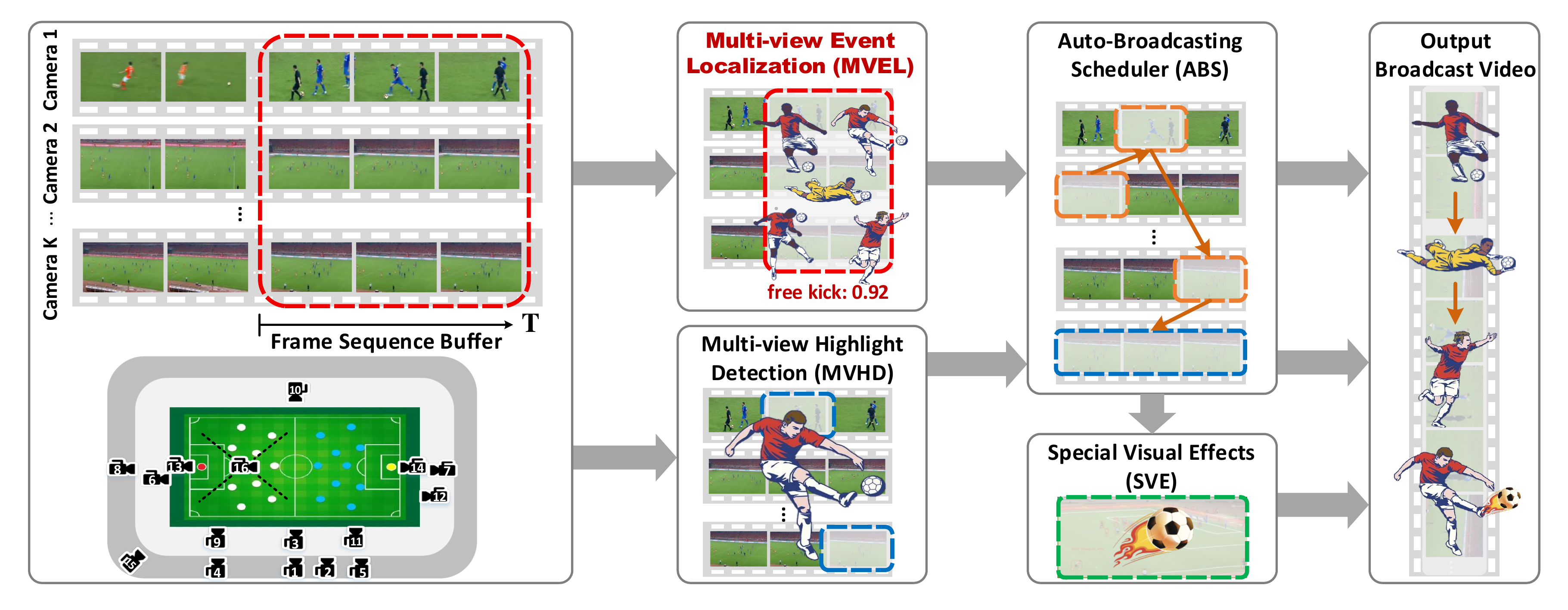}
    \vspace{-0.2in}
	\caption{\small An overview of our Smart Director system. The inputs of our system are multiple synchronous streaming videos from $K$ cameras. By simultaneously processing the multi-view videos with temporal sliding buffers of size $T$, our system firstly feeds the buffers into the MVEL module to detect the events of interest. Within the detected event duration, the MVHD module will select the most attractive views as highlights for potential replays. Having the outputs of previous modules, ABS will automatically produce the broadcast videos through camera view selection, and/or slow-motion replays for significant events such as free-kick and Special Visual Effects video editing for some specific events.}
\label{fig2}
\vspace{-0.1in}
\end{figure*}

To this end, we first need to understand and master the nature of sports broadcast. Goldlust \cite{barnfield2013soccer,goldlust1987playing} suggests the broadcast of sports is constructed of three simultaneously occurring events: the sport event, stadium event, and medium event, where the sport event is defined as the actions on the field as well as related activities taking place on the sidelines. As a storyteller, the director should understand the action flow of the sport and anticipate the action point \cite{owens2015television,zuo2020idirector}, thus captures the critical actions to present the viewer a comprehensive story. Hence, we argue that the broadcast directing is basically an event-driven process. Our argument is also consistent with the observation that human directors usually trigger broadcast changes such as view switches and replays when some critical actions are taking place. Following this line of thought, we develop an end-to-end event-driven directing system for live sports broadcasting (i.e., soccer games in this paper). To our knowledge, there is no such comprehensive broadcasting system developed in the past. Although a few systems have been proposed to solve the specific problem of camera selection \cite{wang2008automatic,choi2009automatic,daniyal2011multi,wang2014context,ada,chen2018camera}, they can't be counted as an end-to-end directing system. In other words, camera selection is indeed important for multi-view broadcasting, but it is only one part of the entire directing system. Additionally, these systems are usually built on various trivial rules or criteria, such as view clearness \cite{wang2008automatic}, the visibility of objects of interest \cite{choi2009automatic,daniyal2011multi,ada}, and smoothness \cite{chen2018camera}. In contrast, ours is driven by the actions of the sports without varied predefined explicit rules.

As shown in Figure \ref{fig2}, our directing system, called Smart Director, consists of the following modules: Multi-view Event Localization (MVEL), Multi-view Highlight Detection (MVHD), Auto-Broadcasting Scheduler (ABS), and Special Visual Effects (SVE). The MVEL module is trained to master the action flow of a sport game by temporally localizing events of interest. It basically prompts a series of directing operations like camera switches and slow-motion replays. Exploiting the multi-camera characteristics of our system,  we specifically design a novel multi-view convolutional network linked by multi-view relation blocks for event detection. As a result, our MVEL outperforms the state-of-the-art event detection approaches \cite{lin2017single,gao2018ctap}, which are not invented to deal with this new problem of multi-view event detection. As we know, each camera in our system has its own ``special assignment'', and its importance to be highlighted as replays in a broadcast varies as per specific events. For example, the ``close-up camera'' provides the close-ups for the actions and players during an attractive ``player falling'' event, while the ``goal-line camera'' offers an unobstructed view of the entire goal area, which is an important view for broadcasting the ``goal kick'' event. Hence, the MVHD module is trained to predict the highlight scores of an event at different views for better slow-motion replays~generation.

Driven by aforementioned event modules, the learning-based scheduler ABS is invented to determine how to deliver the final broadcasting story. In order to capture the viewer's emotions, a story should contains three parts: the beginning, the middle, and the end \cite{owens2015television}. Hence, our scheduler is designed to select clips for ``event-in-progress'', ``event-begin'' and ``event-end'' to composite an event story.
Learning from demonstrations via multi-layer perception classifiers and integer programming, the scheduler is able to figure out an optimal solution to produce the final entertaining on-air videos. Unlike previous camera selection systems \cite{choi2009automatic,daniyal2011multi,wang2014context,chen2018camera}, which usually follow one or two specific rules to select a view, our scheduler leverages the comprehensive knowledge regarding an event, such as its content, duration, priority and the shot's perceptual quality, to generate the critical moments. Our learning-based approach also enables the possibility to produce personalized broadcast videos by learning from personal preferences.

Our Smart Director has been thoroughly evaluated on a 99-hour soccer video dataset collected from six matches of the Chinese Football Association Super League. Both objective and subjective experiments have demonstrated that our system can produce broadcast videos with comparable quality to the ``ground-truth'' videos directed by humans (e.g., 0.618 v.s. 0.658 in terms of subjective satisfaction). Additionally, we have evaluated each individual module. For example, MVEL can outperform the state-of-the-art detection approaches by 2.5\% in terms of mean average precision.

In summary, we have made the following contributions in our system:
\textbf{(I).} To our best knowledge, the Smart Director is the first real event-driven directing system for live sports broadcasting~with comprehensive functionalities mimicking human~directors.
\textbf{(II).} We solve a novel problem of multi-view event localization by detecting events of interest using multi-view relation network blocks, outperforming the state-of-the-art single-view event detection methods.
\textbf{(III).} We propose a learning-based auto-broadcasting scheduler determining what and how to present a sport event via operations like view selections and slow-motion replays. Unlike the rule-based camera selection, our scheduler is fully driven by sport events to produce broadcast videos.

\section{Related Work}

Camera selection is an important part of broadcast directing (i.e., video editing), where the goal is to choose optimal camera views for conveying a story based on various criteria, such as the visibility of objects-of-interest \cite{choi2009automatic,wang2008automatic,daniyal2011multi,wang2014context,li2019psdirector}, smoothness \cite{chen2018camera,chen2019learning}, ``gravity'' (i.e., layout) of players \cite{lefevre2018automatic}, and so on. Most such video editing systems are built on the predefined rules-of-thumb \cite{choi2009automatic,wang2008automatic,daniyal2011multi,wang2014context,leake2017computational}, which are conveyed through computational measurements like ball size, number of subjects, and speaker visible. However, it is cumbersome or infeasible to make every single rule computable, which limits the applications of rule-based approaches. To solve this issue, in our system we design a learning-based scheduler to learn human directing styles from training videos (i.e., the ones produced by human directors). Basically, this data-driven mechanism implicitly learns the underlying ``rules'' of broadcasting from videos, and thus it avoids defining various rules and handcrafting their corresponding features. Our scheduler also differs from other learning-based approaches \cite{ada,chen2018camera,chen2019learning}. Instead of learning an explicit single objective like ``smoothness'' \cite{chen2018camera,chen2019learning} and ``player distribution'' \cite{ada}, our scheduler is driven by sport events' characteristics.

Event detection and highlight extraction are important for sports video analysis and broadcasting. Earlier approaches have been well summarized in \cite{shih2017survey}. The task of event/action detection is to temporally localize the target event. Leveraging the deep learning networks, the state-of-the-art approaches detect events via a two-stage process including two consecutive stages of temporal region proposal and action classification \cite{shou2016temporal,Zhao_2017_ICCV,mettes2019pointly}, or a single-stage strategy conducting action proposal and classification simultaneously \cite{lin2017single,long2019gaussian,Buch_2017_CVPR,zhao2019dance}. Most previous approaches, however, are specifically invented for event localization from a single camera view. Event detection from multiple cameras has not been well addressed. To this end, we successfully develop a novel multi-view events localization network consisting of internal multi-view relation blocks to capture the cross-view relations for better event detection.

Highlight detection is generally exploited to emphasize the occurrence of significant events. It is a technique commonly used for  video summarization and slow-motion replay in sports video analysis \cite{shih2017survey}. Early proposed approaches employ various hand-crafted low-level visual features for highlight detection \cite{rui2000automatically,wang2004automatic,javed2016efficient,raventos2015automatic,shih2017survey}. Recently, various deep learning networks like DCNN \cite{yao2016highlight} and LSTM \cite{zhang2016video} have been successfully applied for highlight extraction in general video summarization \cite{kim2018exploiting,xiong2019less,zhang2016video,yao2016highlight}. However, extracting highlights from a multi-camera broadcasting system has not been well explored. Collaborating with the multi-view event detection, we design a multi-view ranking loss to select the highlights from multiple views.

\begin{figure*}[!tb]
\vspace{-0.0in}
	\centering\includegraphics[width=0.96\textwidth]{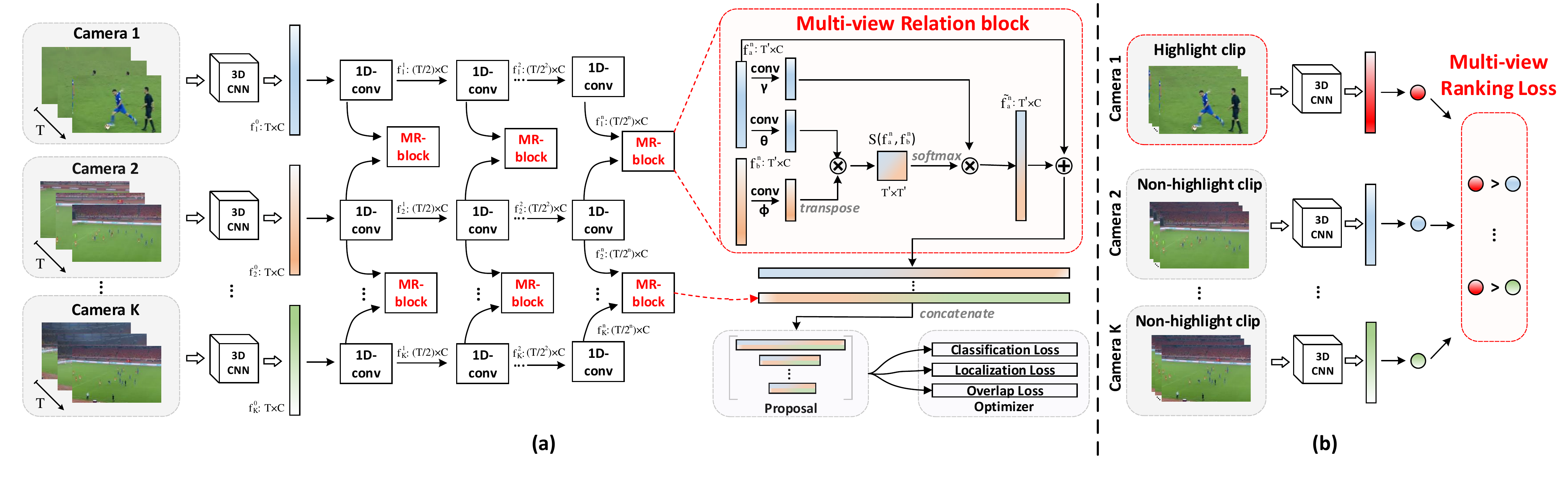}
    \vspace{-0.30in}
	\caption{\small The architectures of (a) Multi-view Event Localization and (b) Multi-view Highlight Detection modules.}
\label{fig2:1}
\vspace{-0.10in}
\end{figure*}

\section{Proposed System}

Figure \ref{fig2} depicts an overview of our system, which is composed of four components: Multi-view Event Localization (MVEL), Multi-view Highlight Detection (MVHD), Auto-Broadcasting Scheduler (ABS), and Special Visual Effects (SVE). Technically, in order to cache sufficient contextual information for video content analysis, we sequentially process the streaming video from each view with a sliding frame sequence buffer (of length $T=30$ \emph{sec}). First, all the recorded frame sequence buffers from $K$ views are fed into the MVEL module to temporally localize and recognize the events of interest. Meanwhile, we adopt the MVHD module to choose the video clips across all $K$ camera views that contain the moments of viewer's major or special interest (i.e., event highlight). These highlights are candidates for the slow-motion replay generation.
After that, the ABS scheduler is trained to produce the final broadcast video conditioned on the recognized events and detected highlights within the multi-view frame sequence buffers. During this production process, the scheduler determines to select the main camera stream or the event-specific video segment with the capability to choose the event beginning, in-progress and end to frame the event story. Note that for some detected high-profile events (e.g., free kick), the scheduler will call the SVE to add visual effects (e.g., highlighted shooting trajectory, shooting distance estimation, and slow motion replay), to increase viewers' engagement.

\vspace{-0.10in}
\subsection{Multi-view Event Localization}
As Smart Director is an event-driven system, we first present the MEVL, its core module which aims to temporally localize and recognize the events of interest from multiple cameras. Although the typical two-stage (``detection by classification'' \cite{Gaidon:PAMI13,Oneata:ICCV13}) and single-stage (``classification and detection jointly'' \cite{lin2017single,li2018jointly}) approaches for temporal event detection have been very successful, they are only applicable to single-view video and lack the capability to exploit the holistic contextual information across views. Taking inspiration from relation modeling in image/video understanding \cite{cai2019exploring,deng2019relation,pan2020x,yao2018exploring,yao2019hierarchy}, we invent a multi-view event localization network to contextually encode the video content from multiple views for event detection via multi-view relation blocks.
As illustrated in Figure \ref{fig2:1} (a), the network architecture of MVEL consists of two stages: the 3D feature extractor and the cascaded 1D temporal convolutional layers with multi-view relation block (MR-Block). Specifically, the 3D feature extractor encodes frame sequence buffers from each view as feature maps, which will be fed into the cascaded 1D convolutional layers to generate multiple proposals in different temporal scales. To further enhance the feature map of each view, the MR-Block is designed to exploit the relations across views. All enhanced feature maps are finally aggregated via the channel-wise concatenation into the holistic feature representation for proposal generation.

\textbf{Base 3D Feature Extractor.}
Given the video buffers from $K$ views, we first extract clip-level features from continuous clips within each video buffer via the P3D extractor \cite{qiu2017learning}, which can capture both appearance and motion information of the video. Please note that the motion information here reflects not only the player motion in the video buffers, but also the camera motion (i.e., the camera operators' reactions).
Specifically, we denote the $\{f_{k,i}\}_{i=0}^{T-1}$ as the 3D feature sequence extracted from the video buffer of $k$-th view. We concatenate all the 3D features of each view into one feature map, which will be fed into eight 1D temporal convolutional layers (convs) for temporal event proposal generation at each temporal scale.

\textbf{Cascaded 1D Convs with Multi-view Relation Block.}
Considering that the video content from different views naturally conveys complementary cues to depict the same event, we upgrade the 1D temporal convs with a novel multi-view relation block to additionally capture the relations between two views. The multi-view video content is thus contextually encoded with the relations across views, leading to the further boost of event localization.
Formally, let $f_a^n$ and $f_b^n$ denote the feature map of the $n$-th 1D convolutional layer from the $a$-th and $b$-th view, respectively. The feature dimension of $f_a^n$ and $f_b^n$ is $T'\times C$. The multi-view relation block firstly measures the relation between $i$-th and $j$-th temporal position of $f_a^n$ and $f_b^n$ via dot-product similarity:
\begin{equation}\label{Eq2:1}
\begin{split}
&S(f_{a,i}^n,f_{b,j}^n) = \theta(f_{a,i}^n)\phi(f_{b,j}^n)^\top, ~~~~~~~i,j \in \{0,1,...,T'-1\},
\end{split}
\end{equation}
where $\theta$ and $\phi$ represents two feature embedding functions. $S(f_{a,i}^n,f_{b,j}^n)$ denotes one entry of similarity matrix $S$.
Next, we normalize the similarity matrix $S$ by row through a softmax operation to obtain the attention map. The attended feature embedding of $i$-th position in $a$-th view is thus calculated by aggregating the features of all temporal positions in $a$-th view with the relations between $a$-th and $b$-th views:
\begin{equation}\label{Eq2:2}
\begin{split}
&\tilde{f}_{a,i}^n = \sum\nolimits_{j=0}^{T'-1}{\gamma(f_{a,j}^n) S(f_{a,i}^n,f_{b,j}^n)},
\end{split}
\end{equation}
where $\gamma$ is the embedding function. Besides, a residual connection is additionally constructed between input and output of multi-view relation block, making the optimization easier.
Accordingly, among all $K$ views, we can obtain the $K(K-1)$ attended feature maps by exploring the relations between every two views. The holistic video representation is produced by concatenating all~the feature maps in a channel-wise manner. The $t$-th cell in this holistic feature map corresponds to an event proposal at the $n$-th temporal scale, whose default center location $a_c$ and width $a_w$ are defined as
\begin{equation}\label{Eq2:3}
a_c = (t+0.5)/T',~~~~a_w = r_d /T',
\end{equation}
where $r_d$ is the temporal scale ratio.

\textbf{Optimization and Inference.}
For each cell (i.e., proposal) in the holistic feature map, three 1D convolutional layers are separately employed to generate event classification scores, localization offsets and overlap scores, respectively. In particular, the classification scores $\mathbf{s^e}=[s^e_0, s^e_1, ..., s^e_{C}]$ indicate the probabilities over $C$ event classes plus one ``background" class. The vector $(\Delta c, \Delta w)$ denotes the temporal offsets to the default proposal $(a_c,a_w)$ (i.e., center location and width). The coordinates of each proposal are thus refined as:
\begin{equation}\label{Eq2:4}
       \begin{split}
              \varphi_c = a_c + \alpha_1 a_w \Delta c, ~~~\varphi_w = a_w e^{\alpha_2 \Delta w}~,
       \end{split}
\end{equation}
where \mbox{$\varphi_c$} and \mbox{$\varphi_w$} are adjusted center location and width of the proposal. $\alpha _1$ and $\alpha _2$ ($\alpha _1=\alpha _2=0.1$) are used to control the impact of temporal offsets. Moreover, we measure the overlap score $s_{op}$ to represent the precise Intersection over Union (IoU) prediction of the proposal, which will benefit proposal re-ranking. At training stage, we accumulate all proposals from the holistic feature at each temporal scale for optimization. The overall objective function is formulated as a multi-task loss by integrating event classification loss and two regression losses (i.e., localization loss and overlap loss). During inference, the ranking score $s_f$ of each proposal depends on both classification $\mathbf{s^e}$ and overlap $s_{op}$: $s_f = \max(\mathbf{s^e}) \cdot s_{op}$. Given all predicted event instances, we utilize the non-maximum suppression (NMS) to remove high overlap event instances in post-processing.

\subsection{Multi-view Highlight Detection}
In computational sports video generation, highlight detection is generally used to create slow-motion replay. Our MVEL can detect candidate events for highlights, but it is unable to tell which camera view is good for highlight replay. To overcome this problem, we innovate the MVHD module to assign highlight scores all views of an event. The higher the score, the more chance the view (i.e., video clip) to be selected for highlight. Inspired by the relative relation modeling via ranking \cite{li2019deep,li2019learning,pan2014click,pan2016learning,pan2015semi,yao2016highlight,seco}, one way to train the MVHD is to utilize the pairwise ranking objective that enforces the score of highlighted clip higher than that of non-highlighted clip. Nevertheless, this will result in a sub-optimal solution, because the pairwise ranking objective can only compare one pair of highlighted and non-highlighted clips, leaving other non-highlighted views unexploited. In contrast, we design a multi-view ranking loss that optimizes the highlight detection by pursuing a higher score for a highlighted clip against all the non-highlighted ones, as shown in Figure \ref{fig2:1} (b).

Formally, suppose we have a set of video clips $\{v^t_+, v^t_1,v^t_2,...,v^t_{K-1}\}$ corresponding to the $K$ camera views at time step $t$, where $v^t_+$ denotes the highlight and $\{v^t_i\}^{K-1}_{i=1}$ are non-highlights. All video clips are separately fed into $K$ identical highlight detection modules with shared 3D CNN based architecture. Let $\{h(v^t_+), h(v^t_1),h(v^t_2),...,h(v^t_{K-1})\}$ be the highlight scores for all clips, the multi-view ranking loss is then measured as:
\begin{equation}\label{Eq3}
       \small
       {L}_{ranking} = \sum\nolimits_{i=0}^{K-1}{max(0, 1-h(v^t_+)+h(v^t_i))}.
\end{equation}
Accordingly, by minimizing the multi-view ranking loss, the score of the highlighted clip will be enforced to be higher than that of all the non-highlighted ones.

\subsection{Auto-Broadcasting Scheduler}

Given the localized event and highlighted scores for all views, how to capitalize on them for the final broadcast video generation? To this end, we invent the Auto-Broadcasting Scheduler (ABS), which determines what content (e.g., slow-motion replay and camera view, etc.) to be on-air.
Among all $K$ cameras, the primary one (i.e., ``main camera'') is positioned on Television Gantry exactly along the halfway line. It provides the main wide-shot coverage of the match. The scenes in the manually generated broadcast videos are predominantly derived from this main camera. Therefore, if no event is detected in the multi-view video buffers, our ABS directly chooses the main camera for broadcasting. Otherwise, for each detected event, ABS will blend the clips from different views to compose an event story for broadcasting as follows.

\begin{figure*}[!tb]
\vspace{-0.0in}
	\centering\includegraphics[width=0.76\textwidth]{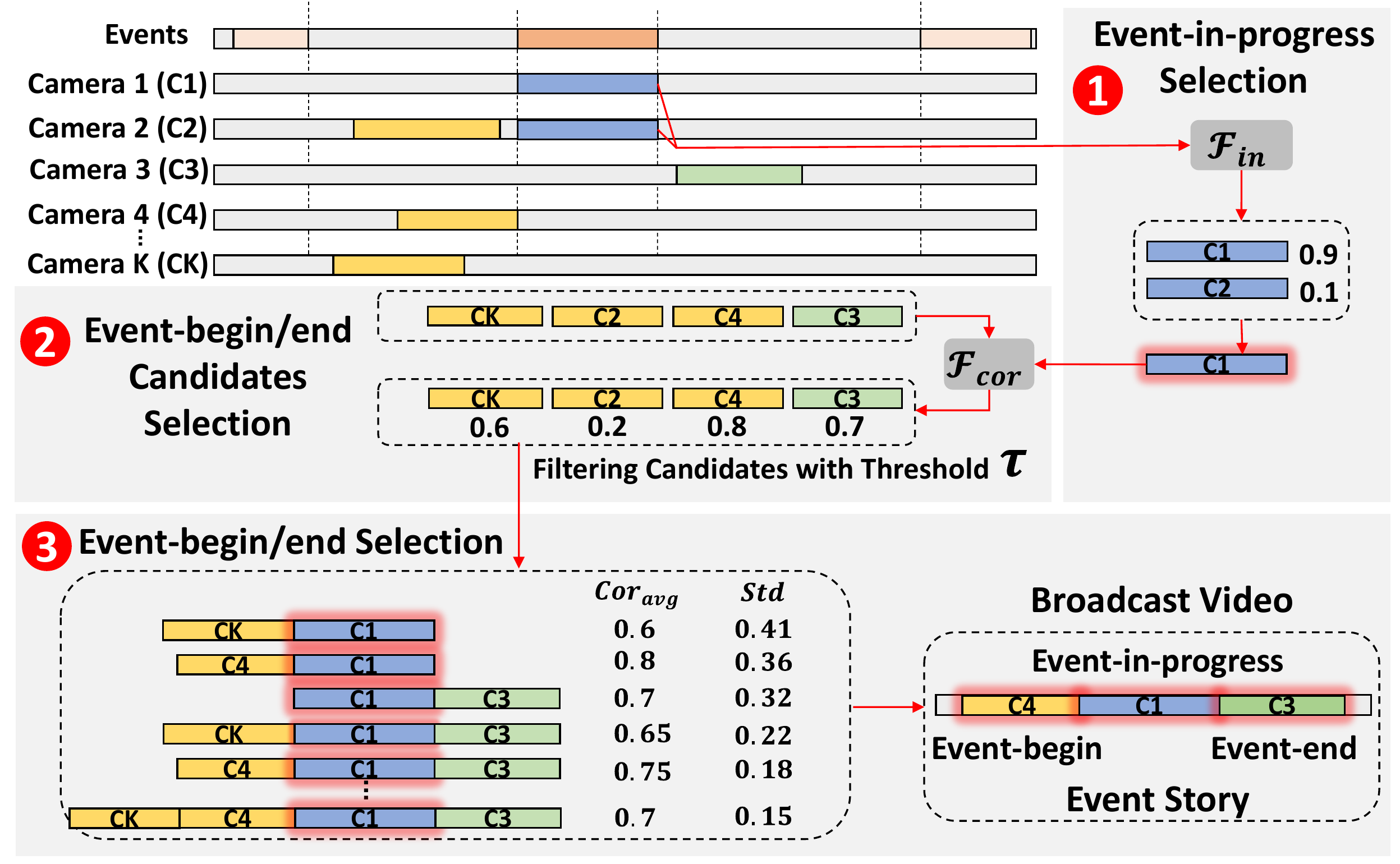}
    \vspace{-0.1in}
	\caption{\small The ABS's event video generation pipeline. Given a detected event, \textcircled{1} the scheduler uses a view classifier $\mathcal{F}_{in}$ to determine the view for the event-in-progress. \textcircled{2} Having the selected event-in-progress clip, it utilizes a binary correlation classifier $\mathcal{F}_{cor}$ to create a group of highly correlated event-begin/end candidates. \textcircled{3} The scheduler chooses the final event-begin/end clips from all candidates by solving an integer programming problem that simultaneously maximizes the holistic correlation and camera view~diversity. }
\label{figABS}
\vspace{-0.10in}
\end{figure*}

Formally, let $\textbf{E} = \{E_1, E_2, ... , E_N\}$ be a sequence of $N$ events detected within sequence buffer of $T$ \emph{sec}, and each event $E_i = (p, t_i^s, t_i^e)$, where $p$ is the event priority level, $t_i^s$ and  $t_i^e$  are the start and end time of this event, respectively. In this paper, we set $p$ as 1 for the four important events (i.e., ``shooting,'' ``player falling,'' ``corner kick,'' and ``free kick''), and $p$ as 0 for the other events. The scheduler creates an event story (i.e., a detailed broadcasting video) for each detected event $E_i$.
Considering the fact that the event durations are varied across different event types and some of them are even hard-understood with ambiguous starting/ending point, it is indeed not trivial to temporally localize the holistic and comprehensive process of each event in a single shot. To alleviate this issue, in ABS, we decompose the holistic process of each event into three video clips from different views, i.e., \textbf{event-begin}, \textbf{event-in-progress}, and \textbf{event-end}, which clearly reflects the beginning, the middle, and the end of the general event story respectively. Note that here the event-in-progress clip refers to the video clip containing key player actions in current detected event $E_i$ with clearly defined starting and ending points. For example, in ``shooting'' event, we define the moment of preparing to shoot and the moment of ball flying out as the corresponding starting and ending points. For ``player falling'' event in fouls, the starting point is defined as the beginning of player falling, and the ending point is the moment of one or multiple players lying on the ground. Thus, it is fairly easier to localize the event-in-progress clip than the localization of the whole event story. Furthermore, once the event-in-progress clip is determined, the selection of event-begin or event-end clips can be formulated as the similarity matching problem, that aims to seek the highly correlated clips from the preceding or subsequent clips with the event-in-progress clip. The rationale behind is to encourage all the three components share the similar content (e.g., the same player), i.e., enforcing the generated event story to be temporally coherent. In this way, we reduce the complexity of event story generation problem by dividing it into three sub-problems: event-in-progress selection, event-begin/end candidates' generation, and event-begin/end selection. The detailed pipeline for event story generation is illustrated in Figure \ref{figABS}.

\textbf{Event-in-progress Selection}. The scheduler will first determine which view of the detected event should be presented, namely, creating the event-in-progress clip. As ``free kick'' occurs, for example, viewers generally prefer to the close-up view, which have better details of the event. We frame the view selection of event-in-progress as a classification problem. A view classifier $\mathcal{F}_{in}$ is thus trained to predict the best camera view for event-in-progress conditioned on the holistic representation $\tilde{v}_i$, duration time $\tilde{t}_i$ and priority level $p$ of that event:
\begin{equation}
    v_{in} = \mathcal{F}_{in}(\tilde{v}_i,\tilde{t}_i,p),
\end{equation}
where $\tilde{v}_i$ is the mean pooling of all $K$ (views) clip-level features extracted by the 3D feature extractor learned in MVEL. The event duration $\tilde{t}_i$ is normalized as: $\frac{t_i^e - t_i^s}{T}$.  $\mathcal{F}_{in}$ is implemented as a two stacked fully connected layer, trained with softmax loss.

\textbf{Event-begin/end Candidates Generation}. Next, given the determined event-in-progress clip, the scheduler seeks a group of highly correlated event-begin/end candidates from the preceding/subsequent clips to better depict the beginning/end of the event-in-progress. The candidates (for both event-begin and event-end) generation is formulated as a binary classification problem (correlated or uncorrelated w.r.t the event). In particular, for each camera view, the scheduler firstly performs face detection \cite{zhang2017faceboxes} and video quality evaluation \cite{pech2000diatom} to seek candidates with high quality and valuable player within the temporal range $[max(0, t_{i-1}^{e}), t_i^{s}]$ or $[t_i^{e},min(t_{i+1}^{s},T)]$. Given a candidate clip $S_k$, the following binary correlation classifier is used to predict its correlation score w.r.t the selected event-in-progress clip:
\begin{equation}
s_k^{r} = \mathcal{F}_{cor}([f_F(\hat{t}_k^s, \hat{t}_k^e), f_F(t_i^s, t_i^e), \tilde{v}_i]),
\end{equation}
where $\hat{t}_k^{s}$ and $\hat{t}_k^e$ are the start and end time of candidate clip $S_k$, and $f_F(t^s,t^e)$ is the face representation extracted from a segment ($t^s, t^e$) by a pre-trained face detector. The binary classifier $\mathcal{F}_{cor}$ consists of three fully connected layers, trained with a binary classification loss (correlated/uncorrelated). The correlation score $s_k^{r}$ measures the probability of being ``correlated". Finally, we filter out all candidates with lower correlation scores (less than threshold $\tau$).

\textbf{Event-begin/end Selection}. In this stage, given all event-begin/end candidates $\textbf{S} = \{S_k\}$, the target is to choose an appropriate pair as the final event-begin and event-end clips for the broadcast video. In general, a high-quality event story is expected to be both temporally coherent and representative of the detected event, so all selected event-begin/end clips should be maximally correlated with the event-in-progress. Meanwhile, the broadcast video should be diverse as much as possible to broadcast soccer match from different camera views (except for the main camera).
As a consequence, we frame the selection of event-begin/end clips as an integer programming problem, that aims to simultaneously maximize the holistic correlation and camera view diversity via an optimal solution of event-begin/end selection $\textbf{S}^\star$:
\begin{equation}
\textbf{S}^\star = \arg \mathop {\max }\limits_{\textbf{S}^* \subseteq \textbf{S}} {Cor_{avg}(\textbf{S}^*)-Std(\textbf{S}^*)},
\end{equation}
where $\textbf{S}^*$ denotes a possible group of event-begin/end candidates by aggregating all selected candidates in chronological order. $Cor_{avg}(\textbf{S}^*)$ assesses the holistic correlation of selected event-begin/end candidates by averaging the correlation scores of all candidates within $\textbf{S}^*$. To measure the degree of camera view diversity, we introduce a camera view count vector $\textbf{C}$ to keep track of how many times each camera view (except for the main camera) has been selected for broadcasting till now. Thus, we interpret the camera view diversity over broadcast video with current event-begin/end selection $\textbf{S}^*$ as the standard deviation of the updated $\textbf{C}$ (i.e., $Std(\textbf{S}^*)$). The less the standard deviation of $\textbf{C}$, the higher the camera view diversity. Note that here we normalize the standard deviation into the range of [0,1].

Finally, the scheduler generates the event story video for event $E_i$ by blending the selected event-begin/end clips with the event-in-progress clip. In addition, according to the event type and priority level, the scheduler further decides if any visual effects should be added to the generated video, or any slow-motion replay should be inserted from the highlight detection results of MVHD.
Please note that our Smart Director can safely insert such slow-motion replay into the broadcast video with 30 seconds looking ahead, and thus does not require the real-time anticipation of what may be happening live when inserting replays. Specifically, for the slow-motion replays of important events with high priority level, we directly insert each replay after the event-end clips of that event. If the slow-motion replay occurs to temporally overlap with the following important event, the scheduler will choose the broadcasting of the important event.
After the production of event story video for each detected event, the final output broadcast video is generated by aggregating them in chronological order. Note that if the event stories of two events temporally overlap, the scheduler will preserve the event with higher priority level.

\subsection{Video Editing with Visual Effects}

According to the specific event (e.g., free kick) , the system may select some visual effects as follows making viewers more engaged.

\textbf{Shooting Trajectory Highlight}. The success of ball tracking is the guarantee for this ``shooting trajectory highlight'' visual effect. To this end, we utilize a two-stream deep architecture \cite{zhang2020robust} as our ball tracker. This tracker leverages both appearance and motion cues for object tracking, such that it is more robust to occlusion, motion blur and confusing background. In addition, we adaptively initialize the ball tracker with both optical-flow and appearance based ball detection to deal with deformed soccer ball detection due to high speed object motion.
This visual effect attempts to visualize how the soccer ball travelling after the kicking on the current frame, which can increase the visibility of ball during some critical events.
Figure~\ref{fig:freekick_ball_track_traject} (a) depicts this situation at a free kick. The soccer ball is successfully tracked and the travel trajectory is drawn on the current frame based on the collective ball positions of the current and past frames.

\textbf{Shooting Distance Estimation}. The estimation of shooting distance is attractive to viewers, especially when some exciting events (e.g., free kick) happen. Nevertheless, due to the existence of camera distortion, it is not trivial to directly predict the distance between the soccer ball and goal in real world based on the recorded video. Here we adopt a camera calibration based pipeline to estimate and visualize the real-world shooting distance. Specifically, given one frame from the main camera, we firstly utilize a pix2pix network \cite{CGAN,chen2019mocycle,CameraCalibration} to synthesize the segmentation result of grassland and the detection result of field markings (e.g. filed lines/circles, goal lines). Meanwhile, we perform Canny edge detection and Hough transform \cite{Hough} to enhance the field markings detection results. After that, depending on the detected field makings, the camera calibration method~\cite{CameraCalibration} is applied to project the original input image (with detected ball) on the standard top view field. In this way, the real-world shooting distance is estimated on this standard top view field.
We illustrate the visual effect of shooting distance estimation, shown in Figure~\ref{fig:freekick_ball_track_traject} (b). Given a frame of the detected free kick event moment, our system can identify the ball, the highlighted kicker, the defense wall, and the keeper. With all these reference points, the shooting distance is predicted and visualized to increase the viewers' engagement. All of these auto-detected contents will provide rich clues for free kick analysis, video editing, and other interactive broadcasting.
\begin{figure*}
\begin{tabular}{cc}
        \animategraphics[autoplay,loop,width=0.4\textwidth]{10}{figs/GIF/Trajectory/}{001}{00013}
        \hspace{0.0cm}
        & \hspace{0.0cm}\animategraphics[autoplay,loop,width=0.4\textwidth]{12}{figs/GIF/Distance/}{001}{016}
        \\
        {\small (a) Shooting Trajectory Highlight (GIF Animation)} \hspace{0.0cm} & \hspace{0.0cm} {\small (b) Shooting Distance Estimation (GIF Animation)} \\
    \end{tabular}
    \vspace{-0.15in}
  \caption{Examples of the visual effects at free kicks set plays (PLEASE view the animation in Acrobat Reader): (a) Soccer ball tracking with the ball trajectory marked in soccer-ball-on-fire special effect. (b) Free kick shooting distance estimation.}
  \label{fig:freekick_ball_track_traject}
  \vspace{-0.15in}
\end{figure*}

\section{Experiments}

\subsection{Dataset and System}

To evaluate our system, we collected a large video dataset with 99-hour duration from the Chinese Football Association Super League. It contains six sets of video data from six soccer matches. Each set has $k=10$ views of streaming videos and one manually created broadcast video. In our experiments, we took five matches for training and one match for testing. We also annotated the temporal locations (the starting and ending frame) of six types of events that usually trigger camera view switch, i.e., shooting, player falling, goal kick, throw-in, corner kick, free kick. Their statistics information is detailed in Table~\ref{tab:dataset}. When evaluating the highlight detection, we split each annotated event video into a set of one-second clips evenly sampled from each camera view and select the most interesting one as the highlight.

\begin{table}[!t]
\caption{The statistics of annotated events in our dataset.}
\vspace{-0.1in}
\begin{tabular}{|l|c|c|c|}
    \hline
    \multicolumn{1}{|c|}{Event} & \multicolumn{1}{|c|}{Number} & \multicolumn{2}{c|}{Duration (\emph{sec})} \\\cline{3-4}
    & & Min & Max \\
    \hline
    \hline
    shooting  & 115 & 2.12 & 9.17 \\
    player falling  & 227 & 1.49 & 9.68 \\
    goal kick  & 161 & 1.3 & 8.47 \\
    throw-in  & 103 & 1.26 & 13.03 \\
    corner kick  & 32 & 3.03 & 9.97 \\
    free kick  & 63 & 1.53 & 9.46 \\
    \hline
\end{tabular}
\label{tab:dataset}
\vspace{-0.1in}
\end{table}

In this video dataset, the ground truth broadcast videos of six soccer matches are produced by professional directors from the Chinese Football Association Super League. For each soccer match, the corresponding ground truth broadcast video (consisting of the actual decisions over 10 camera views) is taken by one professional director. Here we perform an additional user study to examine the directing bias among human directors. Specifically, we invite another three professional directors and ask them to manually produce the broadcast videos conditioned on the same input multi-view videos of each soccer match. Then, for each manually directed broadcast video, we measure its camera switch accuracy against the ground truth broadcast video. In other words, we compute the percentage of camera switches that correctly align with ground truth ones. Note that we define a camera switch being correct if the selected view is the same as the ground truth within 1 sec time interval. The final camera switch accuracy score for each ground truth broadcast video is the average of all the three professional directors' scores. As a result, the mean camera switch accuracy score over six ground truth broadcast videos is 92.3\%, which validates the consistency among the directing decisions made by three professional directors and the ground truth ones.

Our system runs on a server with 20 Tesla P40 GPUs. Considering the existence of broadcast~delay, e.g., 30+ \emph{sec} \footnote{\url{https://en.wikipedia.org/wiki/Broadcast_delay}}, in a modern live broadcasting, we set the length of sliding frame sequence buffer $T$ as 30 \emph{sec}, and the sliding step is set as only 1 \emph{sec} to fully capture every event within streaming videos.
During inference, our system sequentially processes the multi-view streaming videos.
Specifically, for each newly cached frame sequence (length: 1 \emph{sec}), the extraction of 3D features in Multi-view Event Localization (MVEL), 3D features in Multi-view Highlight Detection (MVED), and face features in Auto-Broadcasting Scheduler (ABS) takes 0.116, 0.116, and 0.126 \emph{sec}, respectively. Since the feature extraction runs in parallel, the total processing time for newly cached frame sequence is close to 0.126 \emph{sec}. Next, the ABS completes event story generation over the whole frame sequence buffer (length: 30 \emph{sec}) within 0.5 \emph{sec}. Consequently, the overall pipeline finishes in 0.626 \emph{sec}, less than the duration of the newly cached frame sequence, and thus supports sports broadcasting in real time.
In a word, our system is applicable to most broadcasting scenarios with a normal broadcast delay of 30 \emph{sec}. Nevertheless, when the broadcast delay is extremely less than 0.626 \emph{sec}, our system inevitably fails to process the newly cached video buffer before the next buffer comes.

\begin{table}[!tb]
\centering
  \caption{Performance comparison with state-of-the-art single-view action detection methods in our dataset.}
  \vspace{-0.10in}
  \begin{tabular}{|c|c|c|c|c|c|}
    \hline
    Method  & SST \cite{Buch_2017_CVPR}    & SSAD \cite{lin2017single}  & CTAP \cite{gao2018ctap}  &  MVEL$^{-}$ &  MVEL   \\
    \hline\hline
    mAP (\%)     & 47.9 & 50.1 & 53.2  & 54.6 & 55.8  \\\hline
  \end{tabular}
  \label{tab.event}
  \vspace{-0.1in}
\end{table}

\subsection{Evaluation of Event Localization}

We firstly evaluate the capability of event localization in a soccer match. In this experiment, the widely used evaluation metric, i.e., the mean average precision (mAP) with temporal Intersection over Union (IoU) threshold of 0.5, is adopted. Table \ref{tab.event} details the performance comparisons between three state-of-the-art single-view action detection methods and our MVEL. The comparison also includes approach MVEL$^{-}$, a variant of MVEL without multi-view relation blocks. Overall, both MVEL and MVEL$^{-}$ exhibit better performances than other single-view models. In particular, the mAP of MVEL can achieve 55.8\%, making 2.6\% absolute improvement over the best competitor CTAP. The results demonstrate the advantages of exploiting cross-view holistic contextual information for multi-view event localization. In addition, the performance of MVEL$^{-}$ is inferior to that of MVEL, which further confirms the effectiveness of cross-view relations.

\subsection{Evaluation of Highlight Detection}
In this experiment, we evaluate the effectiveness of MVHD for highlighted view detection on the testing soccer match, and report the mAP for all events. We compare the mAP of MVHD with that of three state-of-the-art highlight detection methods (i.e., LSVM, TS-DCNN, re-seq2seq), as shown in Table \ref{tab.highlight} ( four runs have been conducted on our dataset). A clear performance improvement has been achieved by MVHD over other baselines. It verifies the merit of learning highlight detection via multi-view ranking objective, pursing a higher score for the highlighted clip than all the non-highlighted ones from other views.


\begin{table}[!tb]
\centering
  \caption{Performance comparison with several state-of-the-art highlight detection methods in our dataset.}
  \vspace{-0.10in}
  \begin{tabular}{|c|c|c|c|c|}
    \hline
    Method  & LSVM \cite{potapov2014category} & TS-DCNN \cite{yao2016highlight}  & re-seq2seq \cite{zhang2018retrospective}  &  MVHD   \\
    \hline\hline
    mAP (\%)     & 38.7 & 40.2 & 41.9  & 43.1   \\\hline
  \end{tabular}
  \label{tab.highlight}
  \vspace{-0.1in}
\end{table}

\subsection{Evaluation of Auto-Broadcasting Scheduler}

Recall that our scheduler learns a view classifier to select Event-in-progress clip, and a binary correlation classifier to choose Event-begin/Event-end clips. In this subsection, we will evaluate each classifier in our scheduler.

\textbf{Performance of View Classifier}. All 579 event videos that trigger camera view switch in the training matches (by human directors) are used to train the view classifier. It is evaluated on all event videos from the testing match data, where each event has the view selection by a human director. It achieves {93.06\%} of accuracy, demonstrating that our view classifier can substantially mimic human directors in performing camera selection for the generation of event-in-progress clip.

\textbf{Performance of Binary Correlation Classifier}. We train the binary correlation classifier using all clip pairs (i.e., event-begin/end and event-in-progress) in the training match data. Note that the event-begin and event-end selection share one binary correlation classifier. All pairs of event-begin/end and event-in-progress in the testing match data are used for evaluation. As the selection of event-begin/end is basically a problem of ranking clips from all camera views w.r.t their correlation scores, we adopt mAP as the evaluation metric. It turns out the final mAP is 90.36\%, which clearly shows that the binary correlation classifier can select the highly correlated event-begin/end clips for broadcasting.

\textbf{Effect of the Threshold $\tau$}. To clarify the effect of threshold $\tau$ used to filter out the uncorrelated event-begin/end candidates, we compare the results by varying $\tau$ from 0.1 to 0.9. As shown in Table~\ref{tab:tau}, the increase of $\tau$ can lead to performance improvements in terms of Precision, but the Recall drops slightly. As the live broadcasting is sensitive to the appearance of uncorrelated clips in broadcast videos, the Precision is more important than Recall. Accordingly, we set $\tau$ as 0.7, which has the second-best Precision and a better trade-off between all metrics.

\begin{table}[t]
\caption{The effect of threshold $\tau$ in our dataset.}
\vspace{-0.10in}
    \begin{tabular}{|l|c|c|c|}
    \hline
    $\tau$ & Precision (\%)  & Recall (\%) & F1-score \\
    \hline
    \hline
    0.1 & 82.63 & 76.24 & 0.7931 \\
    0.3 & 95.62 & 72.38 & 0.8239 \\
    0.5 & 97.67 & 69.61 & 0.8129 \\
    0.7 & 99.21 & 69.61 & 0.8181 \\
    0.9 & 99.90 & 67.40 & 0.8049 \\
    \hline
    \end{tabular}%
 \label{tab:tau}
 \vspace{-0.1in}
\end{table}

\subsection{Evaluation of Smart Director}

In addition to the evaluations on each individual module, we also conduct both objective and subjective evaluations on the entire broadcasting videos generated by our Smart Director. We compare its performance to two systems: \textbf{Random} and \textbf{Manual}. The \textbf{Random} system, a degraded version of Smart Director, randomly selects views for the event-begin/end and event-in-progress clip. While the \textbf{Manual} system is a human directing system used in current live broadcasting. For the objective evaluation, we directly measure the accuracy of camera switch in our Smart Director against the Manual system. During the subjective evaluations, we invite a total number of 20 viewers with equal gender representation, ages ranging from 23 to 36, to evaluate our production. These viewers have various education backgrounds, including engineering (5), science (5), arts (5), business and others (5). They are soccer enthusiasts, who are familiar with the rules of soccer match and have at least one year of experience watching live soccer broadcasting. We involve two additional tasks that subjectively evaluate the quality of the generated broadcast videos at the entire video-level and individual event-level, respectively.

\begin{figure*}[t]
    \centering
    \vspace{-0.1in}
    \includegraphics[width=0.66\textwidth]{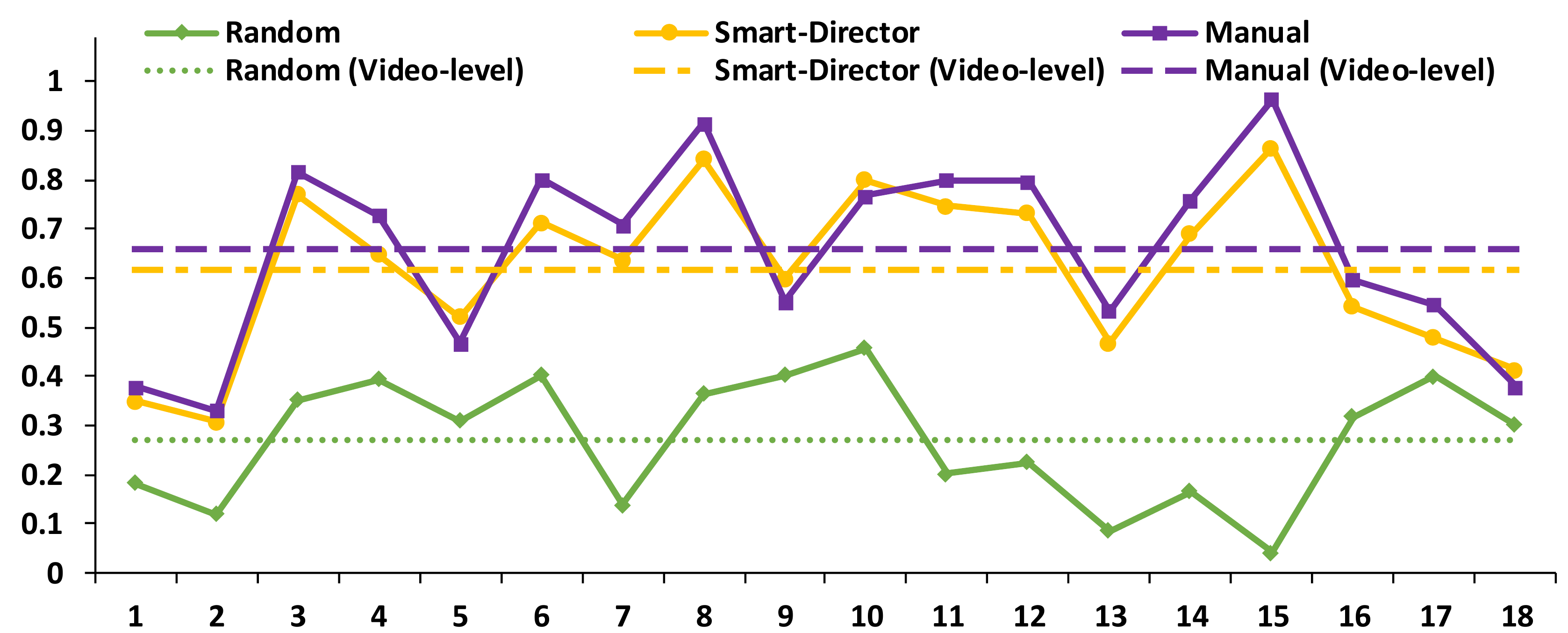}
    \vspace{-0.15in}
    \caption{The video-level subjective evaluation in terms of viewers' satisfactory scores (better viewed in color). For each system, the solid line represents the satisfactory score for each five-minute video segment, and the dotted line denotes the average score over all video segments.}
    \label{fig:sub1}
    \vspace{-0.20in}
\end{figure*}

\textbf{Video-level Objective Evaluation}. Although directing a broadcasting is a subjective job, we still manage to perform an objective evaluation by measuring the accuracy of camera switch in our system against that in the Manual system (i.e., the percentage of camera switches that correctly align with human-directed ones). We define a camera switch being correct if the selected view is
the same as the manually selected within 1 \emph{sec} time interval. As a result, compared to the 212 camera switches made by the Manual system on the testing match, the camera switch accuracy of our system is 81.1\%, which evidences the capability of our system to mimic the human directors for broadcasting. Moreover, when further evaluating ``slow-motion replay'' operations, we observe that our system not only successfully captures all ``slow-motion replay'' operations in Manual system, but also triggers more ``slow-motion replays'' which are missed by the Manual system due to fast-passing events.

\textbf{Video-level Subjective Evaluation}. Given a broadcast video produced by one of the three systems from the 90-minute testing soccer match, we split it evenly into 18 five-minute segments, leading to 54 video segments for all systems. All video segments are mixed and randomly shuffled, then presented to viewers to score among 11 scales of satisfaction ranging from 0 to 10 (0 being the worse and 10 being the best). To eliminate the viewer's bias, we normalize the scores per viewer to $[0,1]$. The final satisfactory score to each segment is the average of all 20 viewers' scores. Figure~\ref{fig:sub1} shows the evaluation results on 18 video segments for each system. The average satisfactory scores for the evaluated systems are: 0.269 (Random), 0.618 (Smart Director), 0.658 (Manual). The large score margin of 0.349 between the Smart Director and Random system clearly shows that our system can produce much meaningful and satisfied broadcast videos. Moreover, the production of our system acquires comparable satisfaction to that of human directors (0.618 vs 0.658 of Smart Director to Manual). Remarkably, the Smart Director even obtains higher scores than the Manual system at four segments (i.e., the 5$^{th}$, 9$^{th}$, 10$^{th}$, and 18$^{th}$ segment). We observe that these segments contain multiple events and the time intervals in between are very short ($\sim$2 \emph{sec}), making it difficult to manually capture the event details for broadcasting. Instead, our system can automatically detect events and select the best camera view to instantly show the event~details.

\begin{figure*}[t]
    \centering
    \includegraphics[width=0.66\textwidth]{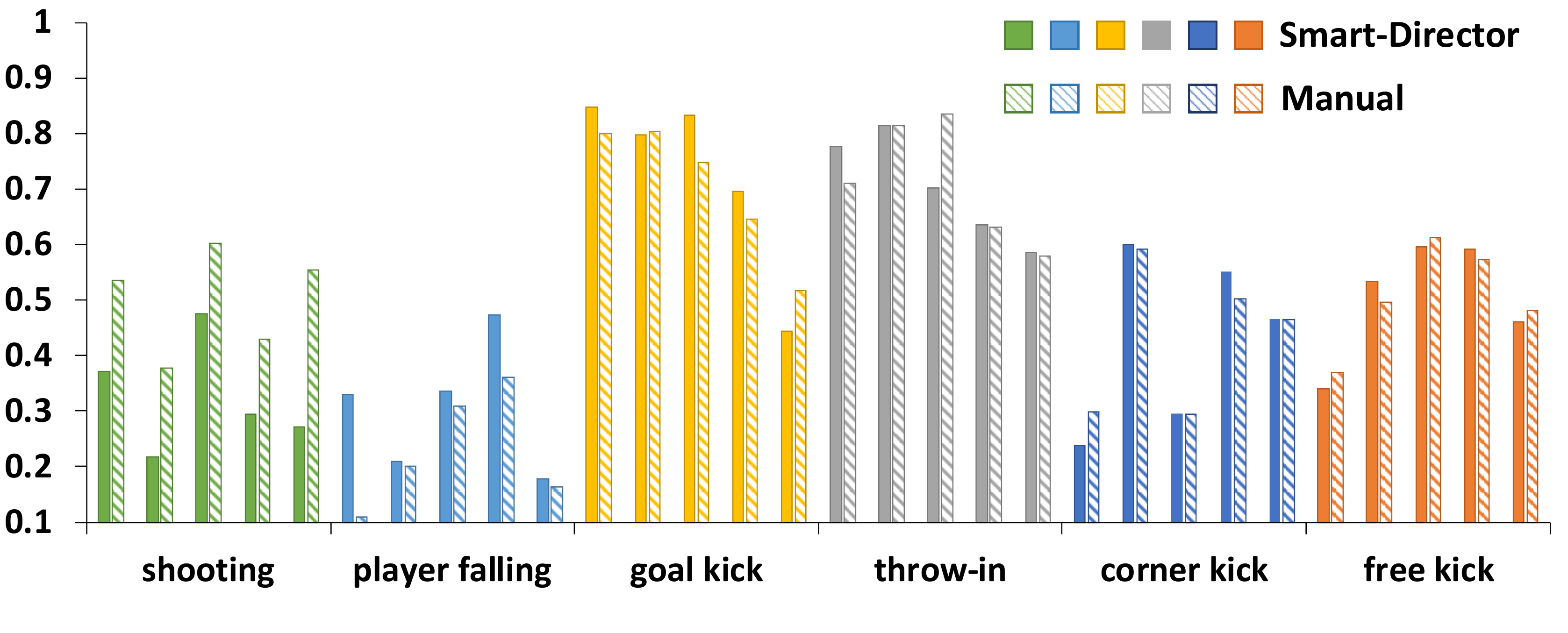}
    \vspace{-0.25in}
    \caption{\small The event-level subjective evaluation in our dataset.}
    \label{fig:sub2}
    \vspace{-0.32in}
\end{figure*}

\textbf{Event-level Subjective Evaluation}. In this evaluation, we will check how well our system works on different event categories. Specifically, we randomly sample five annotated segments from each event category, and ask the viewers to score their satisfaction to the corresponding broadcast video segments generated by different systems. Figure~\ref{fig:sub2} illustrates the event-level subjective evaluation results across different event categories. As we can see, our system acquires similar scores on most event categories (i.e., ``goal kick,'' ``throw-in,'' ``corner kick,'' and ``free kick''). It is worth noting that our system consistently outperforms the Manual system for the five instances in ``player falling''. The results demonstrate the advantage of our system in that it can capture fast-passing events timely and further insert the visual effect of slow-motion replay for broadcasting. Nevertheless, as for the ``shooting'' event, there is a certain gap between our system and the Manual system. We speculate that this is attributed to the exploitation of fine-grained and more focused information (e.g., the positions of goalkeeper and key players, and the actions of key players) when human directors select the best view for ``shooting''. Therefore, incorporating such fine-grained information into the ABS scheduler might strengthen our system, which will be one direction of our future~works.

\section{Conclusions and Future Works}
In this paper, following the concept of treating sports broadcasting as storytelling, we develop Smart Director, an end-to-end event-driven directing system for live sports broadcasting. The system pipeline starts with detecting events of interest from multiple cameras (task of MVEL module) and then compute the highlight scores for multi-views of the event (task of MVHD module). Driven by the understanding to the action flow of the sports game through the first two modules, our scheduler ABS is trained to formulate an output video as an evolution of the event with the beginning, middle and end. As a result, the system's mastery of sport events runs throughout the production of live broadcasting. Smart Director has been tested on real-world soccer datasets, and it has produced high quality broadcast videos which are comparable to the ones generated by human directors.

Our future works are as follows. First, in current version of our Smart Director, the directing system mainly mimics the human-in-the-loop broadcasting process (e.g., camera view selection and slow-motion replays), while ignoring the interaction between directors and camera operators (e.g., instructing a camera operator how to react to major events). The default assumption in our Smart Director is that each camera operator is professionally enough to capture the best views of ongoing game by themselves, that may adversely affect the automatic broadcasting performance especially when some camera operators miss the key moments of significant events. This problem can be alleviated by upgrading our Smart Director with additional action/event forecasting module \cite{kong2017deep,sun2019relational}, that learns to predict the future actions/events for the subsequent streaming video. In this way, our Smart Director can instruct the camera operators to seek the best view or the key player for the upcoming significant event in advance. Moreover, the technique of automatic camera operation can be integrated into the upgraded system with action/event forecasting. Such way leads to a completely automatic directing system, which is composed of both ``smart director'' and ``smart cameras''. Furthermore, we can integrate our Smart Director with a commentator camera to support the broadcasting of game commentators during a timeout or a short break. In particular, the system first localizes the event of ``timeout/shot break'', and then directly selects the commentator camera for broadcasting during this type of event. Note that the ``timeout/shot break'' event has an easily detectable starting point, which is often triggered by referee's gesture. Hence, we can directly employ event localization module over the streaming video from close-up camera (that provides the player and referee close-ups) to detect the ``timeout/shot break'' event.

\begin{acks}
We would like to acknowledge the support from Migu Culture \& Technology Ltd Co. and The Power (Beijing) Sports Ltd Co. for providing the soccer video data in this research.
\end{acks}

\bibliographystyle{ACM-Reference-Format}
\bibliography{smart_director}


\begin{thebibliography}{58}


\ifx \showCODEN    \undefined \def \showCODEN     #1{\unskip}     \fi
\ifx \showDOI      \undefined \def \showDOI       #1{#1}\fi
\ifx \showISBNx    \undefined \def \showISBNx     #1{\unskip}     \fi
\ifx \showISBNxiii \undefined \def \showISBNxiii  #1{\unskip}     \fi
\ifx \showISSN     \undefined \def \showISSN      #1{\unskip}     \fi
\ifx \showLCCN     \undefined \def \showLCCN      #1{\unskip}     \fi
\ifx \shownote     \undefined \def \shownote      #1{#1}          \fi
\ifx \showarticletitle \undefined \def \showarticletitle #1{#1}   \fi
\ifx \showURL      \undefined \def \showURL       {\relax}        \fi
\providecommand\bibfield[2]{#2}
\providecommand\bibinfo[2]{#2}
\providecommand\natexlab[1]{#1}
\providecommand\showeprint[2][]{arXiv:#2}

\bibitem[\protect\citeauthoryear{Barnfield}{Barnfield}{2013}]%
        {barnfield2013soccer}
\bibfield{author}{\bibinfo{person}{Andrew Barnfield}.}
  \bibinfo{year}{2013}\natexlab{}.
\newblock \showarticletitle{Soccer, broadcasting, and narrative: On televising
  a live soccer match}.
\newblock \bibinfo{journal}{\emph{Communication \& Sport}} \bibinfo{volume}{1},
  \bibinfo{number}{4} (\bibinfo{year}{2013}), \bibinfo{pages}{326--341}.
\newblock


\bibitem[\protect\citeauthoryear{Buch, Escorcia, Shen, Ghanem, and
  Carlos~Niebles}{Buch et~al\mbox{.}}{2017}]%
        {Buch_2017_CVPR}
\bibfield{author}{\bibinfo{person}{Shyamal Buch}, \bibinfo{person}{Victor
  Escorcia}, \bibinfo{person}{Chuanqi Shen}, \bibinfo{person}{Bernard Ghanem},
  {and} \bibinfo{person}{Juan Carlos~Niebles}.}
  \bibinfo{year}{2017}\natexlab{}.
\newblock \showarticletitle{Sst: Single-stream temporal action proposals}. In
  \bibinfo{booktitle}{\emph{Proceedings of the IEEE conference on Computer
  Vision and Pattern Recognition}}. \bibinfo{pages}{2911--2920}.
\newblock


\bibitem[\protect\citeauthoryear{Cai, Pan, Ngo, Tian, Duan, and Yao}{Cai
  et~al\mbox{.}}{2019}]%
        {cai2019exploring}
\bibfield{author}{\bibinfo{person}{Qi Cai}, \bibinfo{person}{Yingwei Pan},
  \bibinfo{person}{Chong-Wah Ngo}, \bibinfo{person}{Xinmei Tian},
  \bibinfo{person}{Lingyu Duan}, {and} \bibinfo{person}{Ting Yao}.}
  \bibinfo{year}{2019}\natexlab{}.
\newblock \showarticletitle{Exploring object relation in mean teacher for
  cross-domain detection}. In \bibinfo{booktitle}{\emph{Proceedings of the
  IEEE/CVF Conference on Computer Vision and Pattern Recognition}}.
  \bibinfo{pages}{11457--11466}.
\newblock


\bibitem[\protect\citeauthoryear{Chen, Wang, Heinzle, Carr, Smolic, and
  Gross}{Chen et~al\mbox{.}}{2013}]%
        {ada}
\bibfield{author}{\bibinfo{person}{Christine Chen}, \bibinfo{person}{Oliver
  Wang}, \bibinfo{person}{Simon Heinzle}, \bibinfo{person}{Peter Carr},
  \bibinfo{person}{Aljoscha Smolic}, {and} \bibinfo{person}{Markus Gross}.}
  \bibinfo{year}{2013}\natexlab{}.
\newblock \showarticletitle{Computational sports broadcasting: Automated
  director assistance for live sports}. In \bibinfo{booktitle}{\emph{2013 IEEE
  International Conference on Multimedia and Expo (ICME)}}. IEEE,
  \bibinfo{pages}{1--6}.
\newblock


\bibitem[\protect\citeauthoryear{Chen and Little}{Chen and Little}{2019}]%
        {CameraCalibration}
\bibfield{author}{\bibinfo{person}{Jianhui Chen} {and} \bibinfo{person}{James~J
  Little}.} \bibinfo{year}{2019}\natexlab{}.
\newblock \showarticletitle{Sports camera calibration via synthetic data}. In
  \bibinfo{booktitle}{\emph{Proceedings of the IEEE/CVF Conference on Computer
  Vision and Pattern Recognition Workshops}}. \bibinfo{pages}{0--0}.
\newblock


\bibitem[\protect\citeauthoryear{Chen, Lu, Tian, and Little}{Chen
  et~al\mbox{.}}{2019a}]%
        {chen2019learning}
\bibfield{author}{\bibinfo{person}{Jianhui Chen}, \bibinfo{person}{Keyu Lu},
  \bibinfo{person}{Sijia Tian}, {and} \bibinfo{person}{Jim Little}.}
  \bibinfo{year}{2019}\natexlab{a}.
\newblock \showarticletitle{Learning sports camera selection from internet
  videos}. In \bibinfo{booktitle}{\emph{2019 IEEE Winter Conference on
  Applications of Computer Vision (WACV)}}. IEEE, \bibinfo{pages}{1682--1691}.
\newblock


\bibitem[\protect\citeauthoryear{Chen, Meng, and Little}{Chen
  et~al\mbox{.}}{2018}]%
        {chen2018camera}
\bibfield{author}{\bibinfo{person}{Jianhui Chen}, \bibinfo{person}{Lili Meng},
  {and} \bibinfo{person}{James~J Little}.} \bibinfo{year}{2018}\natexlab{}.
\newblock \showarticletitle{Camera selection for broadcasting soccer games}. In
  \bibinfo{booktitle}{\emph{2018 IEEE Winter Conference on Applications of
  Computer Vision (WACV)}}. IEEE, \bibinfo{pages}{427--435}.
\newblock


\bibitem[\protect\citeauthoryear{Chen, Pan, Yao, Tian, and Mei}{Chen
  et~al\mbox{.}}{2019b}]%
        {chen2019mocycle}
\bibfield{author}{\bibinfo{person}{Yang Chen}, \bibinfo{person}{Yingwei Pan},
  \bibinfo{person}{Ting Yao}, \bibinfo{person}{Xinmei Tian}, {and}
  \bibinfo{person}{Tao Mei}.} \bibinfo{year}{2019}\natexlab{b}.
\newblock \showarticletitle{Mocycle-gan: Unpaired video-to-video translation}.
  In \bibinfo{booktitle}{\emph{Proceedings of the 27th ACM International
  Conference on Multimedia}}. \bibinfo{pages}{647--655}.
\newblock


\bibitem[\protect\citeauthoryear{Choi, Lee, and Seo}{Choi
  et~al\mbox{.}}{2009}]%
        {choi2009automatic}
\bibfield{author}{\bibinfo{person}{Kyu-Hyoung Choi}, \bibinfo{person}{Sang-Wook
  Lee}, {and} \bibinfo{person}{Yong-Duek Seo}.}
  \bibinfo{year}{2009}\natexlab{}.
\newblock \showarticletitle{Automatic broadcast video generation for ball
  sports from multiple views}. In \bibinfo{booktitle}{\emph{Proceedings of the
  Korean Society of Broadcast Engineers Conference}}. The Korean Institute of
  Broadcast and Media Engineers, \bibinfo{pages}{193--198}.
\newblock


\bibitem[\protect\citeauthoryear{Daniyal and Cavallaro}{Daniyal and
  Cavallaro}{2011}]%
        {daniyal2011multi}
\bibfield{author}{\bibinfo{person}{Fahad Daniyal} {and} \bibinfo{person}{Andrea
  Cavallaro}.} \bibinfo{year}{2011}\natexlab{}.
\newblock \showarticletitle{Multi-camera scheduling for video production}. In
  \bibinfo{booktitle}{\emph{2011 Conference for Visual Media Production}}.
  IEEE, \bibinfo{pages}{11--20}.
\newblock


\bibitem[\protect\citeauthoryear{Deng, Pan, Yao, Zhou, Li, and Mei}{Deng
  et~al\mbox{.}}{2019}]%
        {deng2019relation}
\bibfield{author}{\bibinfo{person}{Jiajun Deng}, \bibinfo{person}{Yingwei Pan},
  \bibinfo{person}{Ting Yao}, \bibinfo{person}{Wengang Zhou},
  \bibinfo{person}{Houqiang Li}, {and} \bibinfo{person}{Tao Mei}.}
  \bibinfo{year}{2019}\natexlab{}.
\newblock \showarticletitle{Relation distillation networks for video object
  detection}. In \bibinfo{booktitle}{\emph{Proceedings of the IEEE/CVF
  International Conference on Computer Vision}}. \bibinfo{pages}{7023--7032}.
\newblock


\bibitem[\protect\citeauthoryear{Gaidon, Harchaoui, and Schmid}{Gaidon
  et~al\mbox{.}}{2013}]%
        {Gaidon:PAMI13}
\bibfield{author}{\bibinfo{person}{Adrien Gaidon}, \bibinfo{person}{Zaid
  Harchaoui}, {and} \bibinfo{person}{Cordelia Schmid}.}
  \bibinfo{year}{2013}\natexlab{}.
\newblock \showarticletitle{Temporal localization of actions with actoms}.
\newblock \bibinfo{journal}{\emph{IEEE transactions on pattern analysis and
  machine intelligence}} \bibinfo{volume}{35}, \bibinfo{number}{11}
  (\bibinfo{year}{2013}), \bibinfo{pages}{2782--2795}.
\newblock


\bibitem[\protect\citeauthoryear{Gao, Chen, and Nevatia}{Gao
  et~al\mbox{.}}{2018}]%
        {gao2018ctap}
\bibfield{author}{\bibinfo{person}{Jiyang Gao}, \bibinfo{person}{Kan Chen},
  {and} \bibinfo{person}{Ram Nevatia}.} \bibinfo{year}{2018}\natexlab{}.
\newblock \showarticletitle{Ctap: Complementary temporal action proposal
  generation}. In \bibinfo{booktitle}{\emph{Proceedings of the European
  conference on computer vision (ECCV)}}. \bibinfo{pages}{68--83}.
\newblock


\bibitem[\protect\citeauthoryear{Goldlust}{Goldlust}{2018}]%
        {goldlust1987playing}
\bibfield{author}{\bibinfo{person}{John Goldlust}.}
  \bibinfo{year}{2018}\natexlab{}.
\newblock \bibinfo{booktitle}{\emph{Playing for keeps: Sport, the media and
  society}}.
\newblock \bibinfo{publisher}{Hybrid Publishers}.
\newblock


\bibitem[\protect\citeauthoryear{Isola, Zhu, Zhou, and Efros}{Isola
  et~al\mbox{.}}{2017}]%
        {CGAN}
\bibfield{author}{\bibinfo{person}{Phillip Isola}, \bibinfo{person}{Jun-Yan
  Zhu}, \bibinfo{person}{Tinghui Zhou}, {and} \bibinfo{person}{Alexei~A
  Efros}.} \bibinfo{year}{2017}\natexlab{}.
\newblock \showarticletitle{Image-to-image translation with conditional
  adversarial networks}. In \bibinfo{booktitle}{\emph{Proceedings of the IEEE
  conference on computer vision and pattern recognition}}.
  \bibinfo{pages}{1125--1134}.
\newblock


\bibitem[\protect\citeauthoryear{Javed, Bajwa, Malik, and Irtaza}{Javed
  et~al\mbox{.}}{2016}]%
        {javed2016efficient}
\bibfield{author}{\bibinfo{person}{Ali Javed}, \bibinfo{person}{Khalid~Bashir
  Bajwa}, \bibinfo{person}{Hafiz Malik}, {and} \bibinfo{person}{Aun Irtaza}.}
  \bibinfo{year}{2016}\natexlab{}.
\newblock \showarticletitle{An efficient framework for automatic highlights
  generation from sports videos}.
\newblock \bibinfo{journal}{\emph{IEEE Signal Processing Letters}}
  \bibinfo{volume}{23}, \bibinfo{number}{7} (\bibinfo{year}{2016}),
  \bibinfo{pages}{954--958}.
\newblock


\bibitem[\protect\citeauthoryear{Kim, Mei, Byun, and Yao}{Kim
  et~al\mbox{.}}{2018}]%
        {kim2018exploiting}
\bibfield{author}{\bibinfo{person}{Hoseong Kim}, \bibinfo{person}{Tao Mei},
  \bibinfo{person}{Hyeran Byun}, {and} \bibinfo{person}{Ting Yao}.}
  \bibinfo{year}{2018}\natexlab{}.
\newblock \showarticletitle{Exploiting web images for video highlight detection
  with triplet deep ranking}.
\newblock \bibinfo{journal}{\emph{IEEE Transactions on Multimedia}}
  \bibinfo{volume}{20}, \bibinfo{number}{9} (\bibinfo{year}{2018}),
  \bibinfo{pages}{2415--2426}.
\newblock


\bibitem[\protect\citeauthoryear{Kong, Tao, and Fu}{Kong et~al\mbox{.}}{2017}]%
        {kong2017deep}
\bibfield{author}{\bibinfo{person}{Yu Kong}, \bibinfo{person}{Zhiqiang Tao},
  {and} \bibinfo{person}{Yun Fu}.} \bibinfo{year}{2017}\natexlab{}.
\newblock \showarticletitle{Deep sequential context networks for action
  prediction}. In \bibinfo{booktitle}{\emph{Proceedings of the IEEE conference
  on computer vision and pattern recognition}}. \bibinfo{pages}{1473--1481}.
\newblock


\bibitem[\protect\citeauthoryear{Leake, Davis, Truong, and Agrawala}{Leake
  et~al\mbox{.}}{2017}]%
        {leake2017computational}
\bibfield{author}{\bibinfo{person}{Mackenzie Leake}, \bibinfo{person}{Abe
  Davis}, \bibinfo{person}{Anh Truong}, {and} \bibinfo{person}{Maneesh
  Agrawala}.} \bibinfo{year}{2017}\natexlab{}.
\newblock \showarticletitle{Computational video editing for dialogue-driven
  scenes.}
\newblock \bibinfo{journal}{\emph{ACM Trans. Graph.}} \bibinfo{volume}{36},
  \bibinfo{number}{4} (\bibinfo{year}{2017}), \bibinfo{pages}{130--1}.
\newblock


\bibitem[\protect\citeauthoryear{Lefevre, Bombardier, Charpentier,
  Krommenacker, and Petat}{Lefevre et~al\mbox{.}}{2018}]%
        {lefevre2018automatic}
\bibfield{author}{\bibinfo{person}{Florent Lefevre}, \bibinfo{person}{Vincent
  Bombardier}, \bibinfo{person}{Patrick Charpentier}, \bibinfo{person}{Nicolas
  Krommenacker}, {and} \bibinfo{person}{Bertrand Petat}.}
  \bibinfo{year}{2018}\natexlab{}.
\newblock \showarticletitle{Automatic camera selection in the context of
  basketball game}. In \bibinfo{booktitle}{\emph{International Conference on
  Image and Signal Processing}}. Springer, \bibinfo{pages}{72--79}.
\newblock


\bibitem[\protect\citeauthoryear{Li, Jia, Chen, Gu, and Bao}{Li
  et~al\mbox{.}}{2019a}]%
        {li2019psdirector}
\bibfield{author}{\bibinfo{person}{Chunyang Li}, \bibinfo{person}{Caiyan Jia},
  \bibinfo{person}{Zhineng Chen}, \bibinfo{person}{Xiaoyan Gu}, {and}
  \bibinfo{person}{Hongyun Bao}.} \bibinfo{year}{2019}\natexlab{a}.
\newblock \showarticletitle{psdirector: An automatic director for watching view
  generation from panoramic soccer video}. In
  \bibinfo{booktitle}{\emph{International Conference on Multimedia Modeling}}.
  Springer, \bibinfo{pages}{218--230}.
\newblock


\bibitem[\protect\citeauthoryear{Li, Pan, Yao, Chao, Rui, and Mei}{Li
  et~al\mbox{.}}{2019b}]%
        {li2019learning}
\bibfield{author}{\bibinfo{person}{Yehao Li}, \bibinfo{person}{Yingwei Pan},
  \bibinfo{person}{Ting Yao}, \bibinfo{person}{Hongyang Chao},
  \bibinfo{person}{Yong Rui}, {and} \bibinfo{person}{Tao Mei}.}
  \bibinfo{year}{2019}\natexlab{b}.
\newblock \showarticletitle{Learning click-based deep structure-preserving
  embeddings with visual attention}.
\newblock \bibinfo{journal}{\emph{ACM Transactions on Multimedia Computing,
  Communications, and Applications (TOMM)}} \bibinfo{volume}{15},
  \bibinfo{number}{3} (\bibinfo{year}{2019}), \bibinfo{pages}{1--19}.
\newblock


\bibitem[\protect\citeauthoryear{Li, Yao, Pan, Chao, and Mei}{Li
  et~al\mbox{.}}{2018}]%
        {li2018jointly}
\bibfield{author}{\bibinfo{person}{Yehao Li}, \bibinfo{person}{Ting Yao},
  \bibinfo{person}{Yingwei Pan}, \bibinfo{person}{Hongyang Chao}, {and}
  \bibinfo{person}{Tao Mei}.} \bibinfo{year}{2018}\natexlab{}.
\newblock \showarticletitle{Jointly localizing and describing events for dense
  video captioning}. In \bibinfo{booktitle}{\emph{Proceedings of the IEEE
  Conference on Computer Vision and Pattern Recognition}}.
  \bibinfo{pages}{7492--7500}.
\newblock


\bibitem[\protect\citeauthoryear{Li, Yao, Pan, Chao, and Mei}{Li
  et~al\mbox{.}}{2019c}]%
        {li2019deep}
\bibfield{author}{\bibinfo{person}{Yehao Li}, \bibinfo{person}{Ting Yao},
  \bibinfo{person}{Yingwei Pan}, \bibinfo{person}{Hongyang Chao}, {and}
  \bibinfo{person}{Tao Mei}.} \bibinfo{year}{2019}\natexlab{c}.
\newblock \showarticletitle{Deep metric learning with density adaptivity}.
\newblock \bibinfo{journal}{\emph{IEEE Transactions on Multimedia}}
  \bibinfo{volume}{22}, \bibinfo{number}{5} (\bibinfo{year}{2019}),
  \bibinfo{pages}{1285--1297}.
\newblock


\bibitem[\protect\citeauthoryear{Lin, Zhao, and Shou}{Lin
  et~al\mbox{.}}{2017}]%
        {lin2017single}
\bibfield{author}{\bibinfo{person}{Tianwei Lin}, \bibinfo{person}{Xu Zhao},
  {and} \bibinfo{person}{Zheng Shou}.} \bibinfo{year}{2017}\natexlab{}.
\newblock \showarticletitle{Single shot temporal action detection}. In
  \bibinfo{booktitle}{\emph{Proceedings of the 25th ACM international
  conference on Multimedia}}. \bibinfo{pages}{988--996}.
\newblock


\bibitem[\protect\citeauthoryear{Long, Yao, Qiu, Tian, Luo, and Mei}{Long
  et~al\mbox{.}}{2019}]%
        {long2019gaussian}
\bibfield{author}{\bibinfo{person}{Fuchen Long}, \bibinfo{person}{Ting Yao},
  \bibinfo{person}{Zhaofan Qiu}, \bibinfo{person}{Xinmei Tian},
  \bibinfo{person}{Jiebo Luo}, {and} \bibinfo{person}{Tao Mei}.}
  \bibinfo{year}{2019}\natexlab{}.
\newblock \showarticletitle{Gaussian temporal awareness networks for action
  localization}. In \bibinfo{booktitle}{\emph{Proceedings of the IEEE/CVF
  Conference on Computer Vision and Pattern Recognition}}.
  \bibinfo{pages}{344--353}.
\newblock


\bibitem[\protect\citeauthoryear{Mettes and Snoek}{Mettes and Snoek}{2019}]%
        {mettes2019pointly}
\bibfield{author}{\bibinfo{person}{Pascal Mettes} {and}
  \bibinfo{person}{Cees~GM Snoek}.} \bibinfo{year}{2019}\natexlab{}.
\newblock \showarticletitle{Pointly-supervised action localization}.
\newblock \bibinfo{journal}{\emph{International Journal of Computer Vision}}
  \bibinfo{volume}{127}, \bibinfo{number}{3} (\bibinfo{year}{2019}),
  \bibinfo{pages}{263--281}.
\newblock


\bibitem[\protect\citeauthoryear{Oneata, Verbeek, and Schmid}{Oneata
  et~al\mbox{.}}{2013}]%
        {Oneata:ICCV13}
\bibfield{author}{\bibinfo{person}{Dan Oneata}, \bibinfo{person}{Jakob
  Verbeek}, {and} \bibinfo{person}{Cordelia Schmid}.}
  \bibinfo{year}{2013}\natexlab{}.
\newblock \showarticletitle{Action and event recognition with fisher vectors on
  a compact feature set}. In \bibinfo{booktitle}{\emph{Proceedings of the IEEE
  international conference on computer vision}}. \bibinfo{pages}{1817--1824}.
\newblock


\bibitem[\protect\citeauthoryear{Owens}{Owens}{2015}]%
        {owens2015television}
\bibfield{author}{\bibinfo{person}{Jim Owens}.}
  \bibinfo{year}{2015}\natexlab{}.
\newblock \bibinfo{booktitle}{\emph{Television sports production}}.
\newblock \bibinfo{publisher}{CRC Press}.
\newblock


\bibitem[\protect\citeauthoryear{Pan, Li, Yao, Mei, Li, and Rui}{Pan
  et~al\mbox{.}}{2016}]%
        {pan2016learning}
\bibfield{author}{\bibinfo{person}{Yingwei Pan}, \bibinfo{person}{Yehao Li},
  \bibinfo{person}{Ting Yao}, \bibinfo{person}{Tao Mei},
  \bibinfo{person}{Houqiang Li}, {and} \bibinfo{person}{Yong Rui}.}
  \bibinfo{year}{2016}\natexlab{}.
\newblock \showarticletitle{Learning Deep Intrinsic Video Representation by
  Exploring Temporal Coherence and Graph Structure}. In
  \bibinfo{booktitle}{\emph{IJCAI}}. Citeseer, \bibinfo{pages}{3832--3838}.
\newblock


\bibitem[\protect\citeauthoryear{Pan, Yao, Li, Ngo, and Mei}{Pan
  et~al\mbox{.}}{2015}]%
        {pan2015semi}
\bibfield{author}{\bibinfo{person}{Yingwei Pan}, \bibinfo{person}{Ting Yao},
  \bibinfo{person}{Houqiang Li}, \bibinfo{person}{Chong-Wah Ngo}, {and}
  \bibinfo{person}{Tao Mei}.} \bibinfo{year}{2015}\natexlab{}.
\newblock \showarticletitle{Semi-supervised hashing with semantic confidence
  for large scale visual search}. In \bibinfo{booktitle}{\emph{Proceedings of
  the 38th International ACM SIGIR Conference on Research and Development in
  Information Retrieval}}. \bibinfo{pages}{53--62}.
\newblock


\bibitem[\protect\citeauthoryear{Pan, Yao, Li, and Mei}{Pan
  et~al\mbox{.}}{2020}]%
        {pan2020x}
\bibfield{author}{\bibinfo{person}{Yingwei Pan}, \bibinfo{person}{Ting Yao},
  \bibinfo{person}{Yehao Li}, {and} \bibinfo{person}{Tao Mei}.}
  \bibinfo{year}{2020}\natexlab{}.
\newblock \showarticletitle{X-linear attention networks for image captioning}.
  In \bibinfo{booktitle}{\emph{Proceedings of the IEEE/CVF Conference on
  Computer Vision and Pattern Recognition}}. \bibinfo{pages}{10971--10980}.
\newblock


\bibitem[\protect\citeauthoryear{Pan, Yao, Tian, Li, and Ngo}{Pan
  et~al\mbox{.}}{2014}]%
        {pan2014click}
\bibfield{author}{\bibinfo{person}{Yingwei Pan}, \bibinfo{person}{Ting Yao},
  \bibinfo{person}{Xinmei Tian}, \bibinfo{person}{Houqiang Li}, {and}
  \bibinfo{person}{Chong-Wah Ngo}.} \bibinfo{year}{2014}\natexlab{}.
\newblock \showarticletitle{Click-through-based subspace learning for image
  search}. In \bibinfo{booktitle}{\emph{Proceedings of the 22nd ACM
  international conference on Multimedia}}. \bibinfo{pages}{233--236}.
\newblock


\bibitem[\protect\citeauthoryear{Pech-Pacheco, Crist{\'o}bal,
  Chamorro-Martinez, and Fern{\'a}ndez-Valdivia}{Pech-Pacheco
  et~al\mbox{.}}{2000}]%
        {pech2000diatom}
\bibfield{author}{\bibinfo{person}{Jos{\'e}~Luis Pech-Pacheco},
  \bibinfo{person}{Gabriel Crist{\'o}bal}, \bibinfo{person}{Jes{\'u}s
  Chamorro-Martinez}, {and} \bibinfo{person}{Joaqu{\'\i}n
  Fern{\'a}ndez-Valdivia}.} \bibinfo{year}{2000}\natexlab{}.
\newblock \showarticletitle{Diatom autofocusing in brightfield microscopy: a
  comparative study}. In \bibinfo{booktitle}{\emph{Proceedings 15th
  International Conference on Pattern Recognition. ICPR-2000}},
  Vol.~\bibinfo{volume}{3}. IEEE, \bibinfo{pages}{314--317}.
\newblock


\bibitem[\protect\citeauthoryear{Potapov, Douze, Harchaoui, and Schmid}{Potapov
  et~al\mbox{.}}{2014}]%
        {potapov2014category}
\bibfield{author}{\bibinfo{person}{Danila Potapov}, \bibinfo{person}{Matthijs
  Douze}, \bibinfo{person}{Zaid Harchaoui}, {and} \bibinfo{person}{Cordelia
  Schmid}.} \bibinfo{year}{2014}\natexlab{}.
\newblock \showarticletitle{Category-specific video summarization}. In
  \bibinfo{booktitle}{\emph{European conference on computer vision}}. Springer,
  \bibinfo{pages}{540--555}.
\newblock


\bibitem[\protect\citeauthoryear{Qiu, Yao, and Mei}{Qiu et~al\mbox{.}}{2017}]%
        {qiu2017learning}
\bibfield{author}{\bibinfo{person}{Zhaofan Qiu}, \bibinfo{person}{Ting Yao},
  {and} \bibinfo{person}{Tao Mei}.} \bibinfo{year}{2017}\natexlab{}.
\newblock \showarticletitle{Learning spatio-temporal representation with
  pseudo-3d residual networks}. In \bibinfo{booktitle}{\emph{proceedings of the
  IEEE International Conference on Computer Vision}}.
  \bibinfo{pages}{5533--5541}.
\newblock


\bibitem[\protect\citeauthoryear{Raventos, Quijada, Torres, and
  Tarr{\'e}s}{Raventos et~al\mbox{.}}{2015}]%
        {raventos2015automatic}
\bibfield{author}{\bibinfo{person}{Arnau Raventos}, \bibinfo{person}{Raul
  Quijada}, \bibinfo{person}{Luis Torres}, {and} \bibinfo{person}{Francesc
  Tarr{\'e}s}.} \bibinfo{year}{2015}\natexlab{}.
\newblock \showarticletitle{Automatic summarization of soccer highlights using
  audio-visual descriptors}.
\newblock \bibinfo{journal}{\emph{SpringerPlus}} \bibinfo{volume}{4},
  \bibinfo{number}{1} (\bibinfo{year}{2015}), \bibinfo{pages}{1--19}.
\newblock


\bibitem[\protect\citeauthoryear{Rui, Gupta, and Acero}{Rui
  et~al\mbox{.}}{2000}]%
        {rui2000automatically}
\bibfield{author}{\bibinfo{person}{Yong Rui}, \bibinfo{person}{Anoop Gupta},
  {and} \bibinfo{person}{Alex Acero}.} \bibinfo{year}{2000}\natexlab{}.
\newblock \showarticletitle{Automatically extracting highlights for TV baseball
  programs}. In \bibinfo{booktitle}{\emph{Proceedings of the eighth ACM
  international conference on Multimedia}}. \bibinfo{pages}{105--115}.
\newblock


\bibitem[\protect\citeauthoryear{Shih}{Shih}{2017}]%
        {shih2017survey}
\bibfield{author}{\bibinfo{person}{Huang-Chia Shih}.}
  \bibinfo{year}{2017}\natexlab{}.
\newblock \showarticletitle{A survey of content-aware video analysis for
  sports}.
\newblock \bibinfo{journal}{\emph{IEEE Transactions on Circuits and Systems for
  Video Technology}} \bibinfo{volume}{28}, \bibinfo{number}{5}
  (\bibinfo{year}{2017}), \bibinfo{pages}{1212--1231}.
\newblock


\bibitem[\protect\citeauthoryear{Shou, Wang, and Chang}{Shou
  et~al\mbox{.}}{2016}]%
        {shou2016temporal}
\bibfield{author}{\bibinfo{person}{Zheng Shou}, \bibinfo{person}{Dongang Wang},
  {and} \bibinfo{person}{Shih-Fu Chang}.} \bibinfo{year}{2016}\natexlab{}.
\newblock \showarticletitle{Temporal action localization in untrimmed videos
  via multi-stage cnns}. In \bibinfo{booktitle}{\emph{Proceedings of the IEEE
  conference on computer vision and pattern recognition}}.
  \bibinfo{pages}{1049--1058}.
\newblock


\bibitem[\protect\citeauthoryear{Sun, Shrivastava, Vondrick, Sukthankar,
  Murphy, and Schmid}{Sun et~al\mbox{.}}{2019}]%
        {sun2019relational}
\bibfield{author}{\bibinfo{person}{Chen Sun}, \bibinfo{person}{Abhinav
  Shrivastava}, \bibinfo{person}{Carl Vondrick}, \bibinfo{person}{Rahul
  Sukthankar}, \bibinfo{person}{Kevin Murphy}, {and} \bibinfo{person}{Cordelia
  Schmid}.} \bibinfo{year}{2019}\natexlab{}.
\newblock \showarticletitle{Relational action forecasting}. In
  \bibinfo{booktitle}{\emph{Proceedings of the IEEE/CVF Conference on Computer
  Vision and Pattern Recognition}}. \bibinfo{pages}{273--283}.
\newblock


\bibitem[\protect\citeauthoryear{Wang, Xu, Chng, Lu, and Tian}{Wang
  et~al\mbox{.}}{2008}]%
        {wang2008automatic}
\bibfield{author}{\bibinfo{person}{Jinjun Wang}, \bibinfo{person}{Changsheng
  Xu}, \bibinfo{person}{Engsiong Chng}, \bibinfo{person}{Hanqing Lu}, {and}
  \bibinfo{person}{Qi Tian}.} \bibinfo{year}{2008}\natexlab{}.
\newblock \showarticletitle{Automatic composition of broadcast sports video}.
\newblock \bibinfo{journal}{\emph{Multimedia Systems}} \bibinfo{volume}{14},
  \bibinfo{number}{4} (\bibinfo{year}{2008}), \bibinfo{pages}{179--193}.
\newblock


\bibitem[\protect\citeauthoryear{Wang, Xu, Chng, Wah, and Tian}{Wang
  et~al\mbox{.}}{2004}]%
        {wang2004automatic}
\bibfield{author}{\bibinfo{person}{Jinjun Wang}, \bibinfo{person}{Changsheng
  Xu}, \bibinfo{person}{Engsiong Chng}, \bibinfo{person}{Kongwah Wah}, {and}
  \bibinfo{person}{Qi Tian}.} \bibinfo{year}{2004}\natexlab{}.
\newblock \showarticletitle{Automatic replay generation for soccer video
  broadcasting}. In \bibinfo{booktitle}{\emph{Proceedings of the 12th annual
  ACM international conference on Multimedia}}. \bibinfo{pages}{32--39}.
\newblock


\bibitem[\protect\citeauthoryear{Wang, Hara, Enokibori, Hirayama, and
  Mase}{Wang et~al\mbox{.}}{2016}]%
        {wang2016personal}
\bibfield{author}{\bibinfo{person}{Xueting Wang}, \bibinfo{person}{Kensho
  Hara}, \bibinfo{person}{Yu Enokibori}, \bibinfo{person}{Takatsugu Hirayama},
  {and} \bibinfo{person}{Kenji Mase}.} \bibinfo{year}{2016}\natexlab{}.
\newblock \showarticletitle{Personal multi-view viewpoint recommendation based
  on trajectory distribution of the viewing target}. In
  \bibinfo{booktitle}{\emph{Proceedings of the 24th ACM international
  conference on Multimedia}}. \bibinfo{pages}{471--475}.
\newblock


\bibitem[\protect\citeauthoryear{Wang, Muramatu, Hirayama, and Mase}{Wang
  et~al\mbox{.}}{2014}]%
        {wang2014context}
\bibfield{author}{\bibinfo{person}{Xueting Wang}, \bibinfo{person}{Yuki
  Muramatu}, \bibinfo{person}{Takatsugu Hirayama}, {and} \bibinfo{person}{Kenji
  Mase}.} \bibinfo{year}{2014}\natexlab{}.
\newblock \showarticletitle{Context-dependent viewpoint sequence recommendation
  system for multi-view video}. In \bibinfo{booktitle}{\emph{2014 IEEE
  International Symposium on Multimedia}}. IEEE, \bibinfo{pages}{195--202}.
\newblock


\bibitem[\protect\citeauthoryear{Xiong, Kalantidis, Ghadiyaram, and
  Grauman}{Xiong et~al\mbox{.}}{2019}]%
        {xiong2019less}
\bibfield{author}{\bibinfo{person}{Bo Xiong}, \bibinfo{person}{Yannis
  Kalantidis}, \bibinfo{person}{Deepti Ghadiyaram}, {and}
  \bibinfo{person}{Kristen Grauman}.} \bibinfo{year}{2019}\natexlab{}.
\newblock \showarticletitle{Less is more: Learning highlight detection from
  video duration}. In \bibinfo{booktitle}{\emph{Proceedings of the IEEE/CVF
  Conference on Computer Vision and Pattern Recognition}}.
  \bibinfo{pages}{1258--1267}.
\newblock


\bibitem[\protect\citeauthoryear{Yao, Mei, and Rui}{Yao et~al\mbox{.}}{2016}]%
        {yao2016highlight}
\bibfield{author}{\bibinfo{person}{Ting Yao}, \bibinfo{person}{Tao Mei}, {and}
  \bibinfo{person}{Yong Rui}.} \bibinfo{year}{2016}\natexlab{}.
\newblock \showarticletitle{Highlight detection with pairwise deep ranking for
  first-person video summarization}. In \bibinfo{booktitle}{\emph{CVPR}}.
\newblock


\bibitem[\protect\citeauthoryear{Yao, Pan, Li, and Mei}{Yao
  et~al\mbox{.}}{2018}]%
        {yao2018exploring}
\bibfield{author}{\bibinfo{person}{Ting Yao}, \bibinfo{person}{Yingwei Pan},
  \bibinfo{person}{Yehao Li}, {and} \bibinfo{person}{Tao Mei}.}
  \bibinfo{year}{2018}\natexlab{}.
\newblock \showarticletitle{Exploring visual relationship for image
  captioning}. In \bibinfo{booktitle}{\emph{Proceedings of the European
  conference on computer vision (ECCV)}}. \bibinfo{pages}{684--699}.
\newblock


\bibitem[\protect\citeauthoryear{Yao, Pan, Li, and Mei}{Yao
  et~al\mbox{.}}{2019}]%
        {yao2019hierarchy}
\bibfield{author}{\bibinfo{person}{Ting Yao}, \bibinfo{person}{Yingwei Pan},
  \bibinfo{person}{Yehao Li}, {and} \bibinfo{person}{Tao Mei}.}
  \bibinfo{year}{2019}\natexlab{}.
\newblock \showarticletitle{Hierarchy parsing for image captioning}. In
  \bibinfo{booktitle}{\emph{Proceedings of the IEEE/CVF International
  Conference on Computer Vision}}. \bibinfo{pages}{2621--2629}.
\newblock


\bibitem[\protect\citeauthoryear{Yao, Zhang, Qiu, Pan, and Mei}{Yao
  et~al\mbox{.}}{2020}]%
        {seco}
\bibfield{author}{\bibinfo{person}{Ting Yao}, \bibinfo{person}{Yiheng Zhang},
  \bibinfo{person}{Zhaofan Qiu}, \bibinfo{person}{Yingwei Pan}, {and}
  \bibinfo{person}{Tao Mei}.} \bibinfo{year}{2020}\natexlab{}.
\newblock \showarticletitle{SeCo: Exploring Sequence Supervision for
  Unsupervised Representation Learning}.
\newblock \bibinfo{journal}{\emph{CoRR}}  \bibinfo{volume}{abs/2008.00975}
  (\bibinfo{year}{2020}).
\newblock
\showeprint[arxiv]{2008.00975}
\urldef\tempurl%
\url{https://arxiv.org/abs/2008.00975}
\showURL{%
\tempurl}


\bibitem[\protect\citeauthoryear{Yu and Jain}{Yu and Jain}{1997}]%
        {Hough}
\bibfield{author}{\bibinfo{person}{Bin Yu} {and} \bibinfo{person}{Anil~K
  Jain}.} \bibinfo{year}{1997}\natexlab{}.
\newblock \showarticletitle{Lane boundary detection using a multiresolution
  hough transform}. In \bibinfo{booktitle}{\emph{Proceedings of International
  Conference on Image Processing}}, Vol.~\bibinfo{volume}{2}. IEEE,
  \bibinfo{pages}{748--751}.
\newblock


\bibitem[\protect\citeauthoryear{Zhang, Chao, Sha, and Grauman}{Zhang
  et~al\mbox{.}}{2016}]%
        {zhang2016video}
\bibfield{author}{\bibinfo{person}{Ke Zhang}, \bibinfo{person}{Wei-Lun Chao},
  \bibinfo{person}{Fei Sha}, {and} \bibinfo{person}{Kristen Grauman}.}
  \bibinfo{year}{2016}\natexlab{}.
\newblock \showarticletitle{Video summarization with long short-term memory}.
  In \bibinfo{booktitle}{\emph{European conference on computer vision}}.
  Springer, \bibinfo{pages}{766--782}.
\newblock


\bibitem[\protect\citeauthoryear{Zhang, Grauman, and Sha}{Zhang
  et~al\mbox{.}}{2018}]%
        {zhang2018retrospective}
\bibfield{author}{\bibinfo{person}{Ke Zhang}, \bibinfo{person}{Kristen
  Grauman}, {and} \bibinfo{person}{Fei Sha}.} \bibinfo{year}{2018}\natexlab{}.
\newblock \showarticletitle{Retrospective encoders for video summarization}. In
  \bibinfo{booktitle}{\emph{Proceedings of the European Conference on Computer
  Vision (ECCV)}}. \bibinfo{pages}{383--399}.
\newblock


\bibitem[\protect\citeauthoryear{Zhang, Liu, Wang, Zeng, and Mei}{Zhang
  et~al\mbox{.}}{2020}]%
        {zhang2020robust}
\bibfield{author}{\bibinfo{person}{Ning Zhang}, \bibinfo{person}{Jingen Liu},
  \bibinfo{person}{Ke Wang}, \bibinfo{person}{Dan Zeng}, {and}
  \bibinfo{person}{Tao Mei}.} \bibinfo{year}{2020}\natexlab{}.
\newblock \showarticletitle{Robust Visual Object Tracking with Two-Stream
  Residual Convolutional Networks}.
\newblock \bibinfo{journal}{\emph{CoRR}}  \bibinfo{volume}{abs/2005.06536}
  (\bibinfo{year}{2020}).
\newblock
\showeprint[arxiv]{2005.06536}
\urldef\tempurl%
\url{https://arxiv.org/abs/2005.06536}
\showURL{%
\tempurl}


\bibitem[\protect\citeauthoryear{Zhang, Zhu, Lei, Shi, Wang, and Li}{Zhang
  et~al\mbox{.}}{2017}]%
        {zhang2017faceboxes}
\bibfield{author}{\bibinfo{person}{Shifeng Zhang}, \bibinfo{person}{Xiangyu
  Zhu}, \bibinfo{person}{Zhen Lei}, \bibinfo{person}{Hailin Shi},
  \bibinfo{person}{Xiaobo Wang}, {and} \bibinfo{person}{Stan~Z Li}.}
  \bibinfo{year}{2017}\natexlab{}.
\newblock \showarticletitle{Faceboxes: A CPU real-time face detector with high
  accuracy}. In \bibinfo{booktitle}{\emph{2017 IEEE International Joint
  Conference on Biometrics (IJCB)}}. IEEE, \bibinfo{pages}{1--9}.
\newblock


\bibitem[\protect\citeauthoryear{Zhao and Snoek}{Zhao and Snoek}{2019}]%
        {zhao2019dance}
\bibfield{author}{\bibinfo{person}{Jiaojiao Zhao} {and}
  \bibinfo{person}{Cees~GM Snoek}.} \bibinfo{year}{2019}\natexlab{}.
\newblock \showarticletitle{Dance with flow: Two-in-one stream action
  detection}. In \bibinfo{booktitle}{\emph{Proceedings of the IEEE/CVF
  Conference on Computer Vision and Pattern Recognition}}.
  \bibinfo{pages}{9935--9944}.
\newblock


\bibitem[\protect\citeauthoryear{Zhao, Xiong, Wang, Wu, Tang, and Lin}{Zhao
  et~al\mbox{.}}{2017}]%
        {Zhao_2017_ICCV}
\bibfield{author}{\bibinfo{person}{Yue Zhao}, \bibinfo{person}{Yuanjun Xiong},
  \bibinfo{person}{Limin Wang}, \bibinfo{person}{Zhirong Wu},
  \bibinfo{person}{Xiaoou Tang}, {and} \bibinfo{person}{Dahua Lin}.}
  \bibinfo{year}{2017}\natexlab{}.
\newblock \showarticletitle{Temporal action detection with structured segment
  networks}. In \bibinfo{booktitle}{\emph{Proceedings of the IEEE International
  Conference on Computer Vision}}. \bibinfo{pages}{2914--2923}.
\newblock


\bibitem[\protect\citeauthoryear{Zuo, Chen, Wang, Pan, Yao, Wang, and Mei}{Zuo
  et~al\mbox{.}}{2020}]%
        {zuo2020idirector}
\bibfield{author}{\bibinfo{person}{Jiawei Zuo}, \bibinfo{person}{Yue Chen},
  \bibinfo{person}{Linfang Wang}, \bibinfo{person}{Yingwei Pan},
  \bibinfo{person}{Ting Yao}, \bibinfo{person}{Ke Wang}, {and}
  \bibinfo{person}{Tao Mei}.} \bibinfo{year}{2020}\natexlab{}.
\newblock \showarticletitle{iDirector: An Intelligent Directing System for Live
  Broadcast}. In \bibinfo{booktitle}{\emph{Proceedings of the 28th ACM
  International Conference on Multimedia}}. \bibinfo{pages}{4545--4547}.
\newblock


\end{thebibliography}

\end{document}